\documentclass[a4paper,fleqn]{cas-dc}
\usepackage[numbers]{natbib}
\usepackage{siunitx}
\usepackage[capitalize]{cleveref}
\usepackage{graphicx} 
\usepackage{subcaption}

\newcommand{\etal}{\textit{et al.}}
\newcommand{\Expoverall}{
\begin{figure*}[!ht]
    \centering
    \includegraphics[width=.95\textwidth]{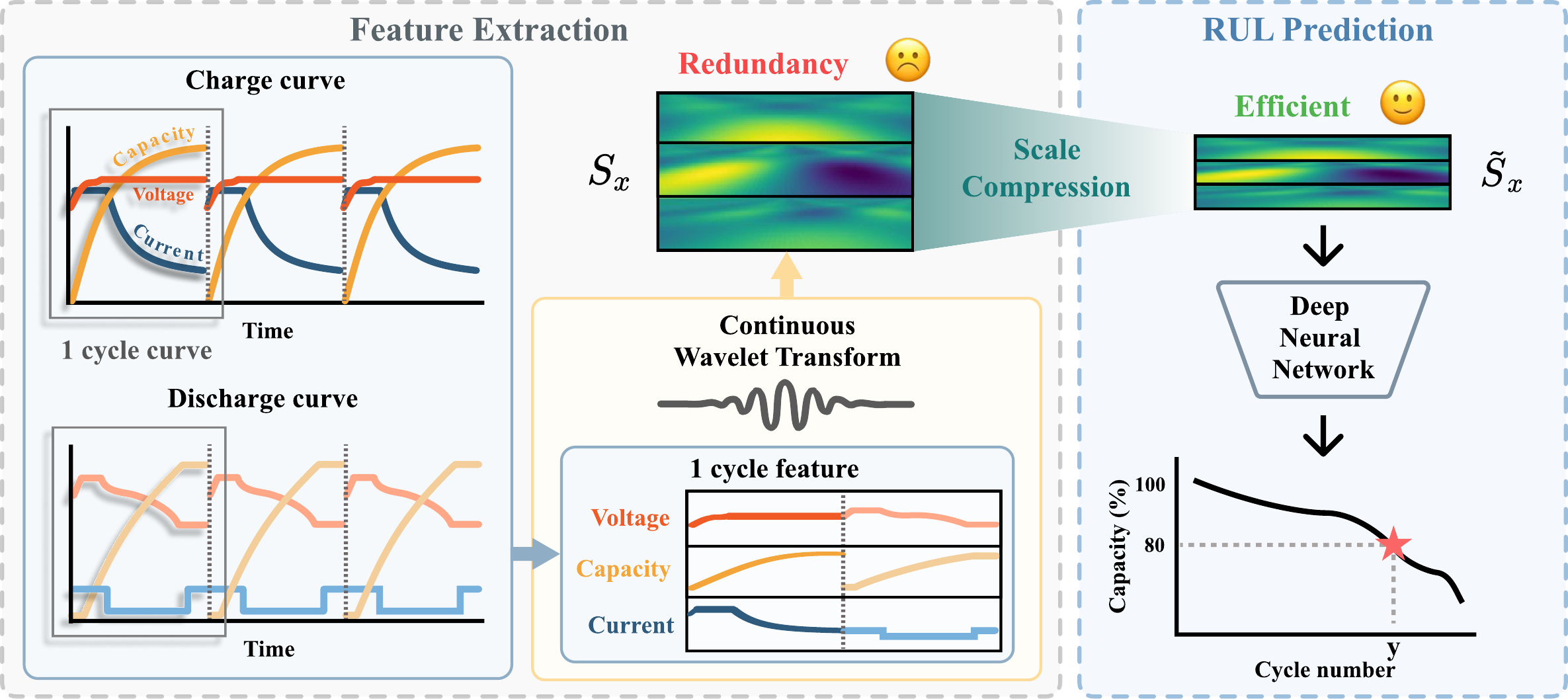}
    \caption{Framework for RUL prediction of lithium-ion batteries. Charge/discharge time-series (voltage, current, capacity) are converted to time-frequency scalograms using the continuous wavelet transform, followed by scale compression to reduce redundancy. The compressed features are fed to a deep neural network for RUL prediction.}
    \label{fig:framework}
\end{figure*}
}

\newcommand{\ExpDatasetFig}{
\begin{figure}[!t]
    \centering
    \begin{subfigure}[t]{0.48\textwidth}
        \centering
        \includegraphics[width=\linewidth]{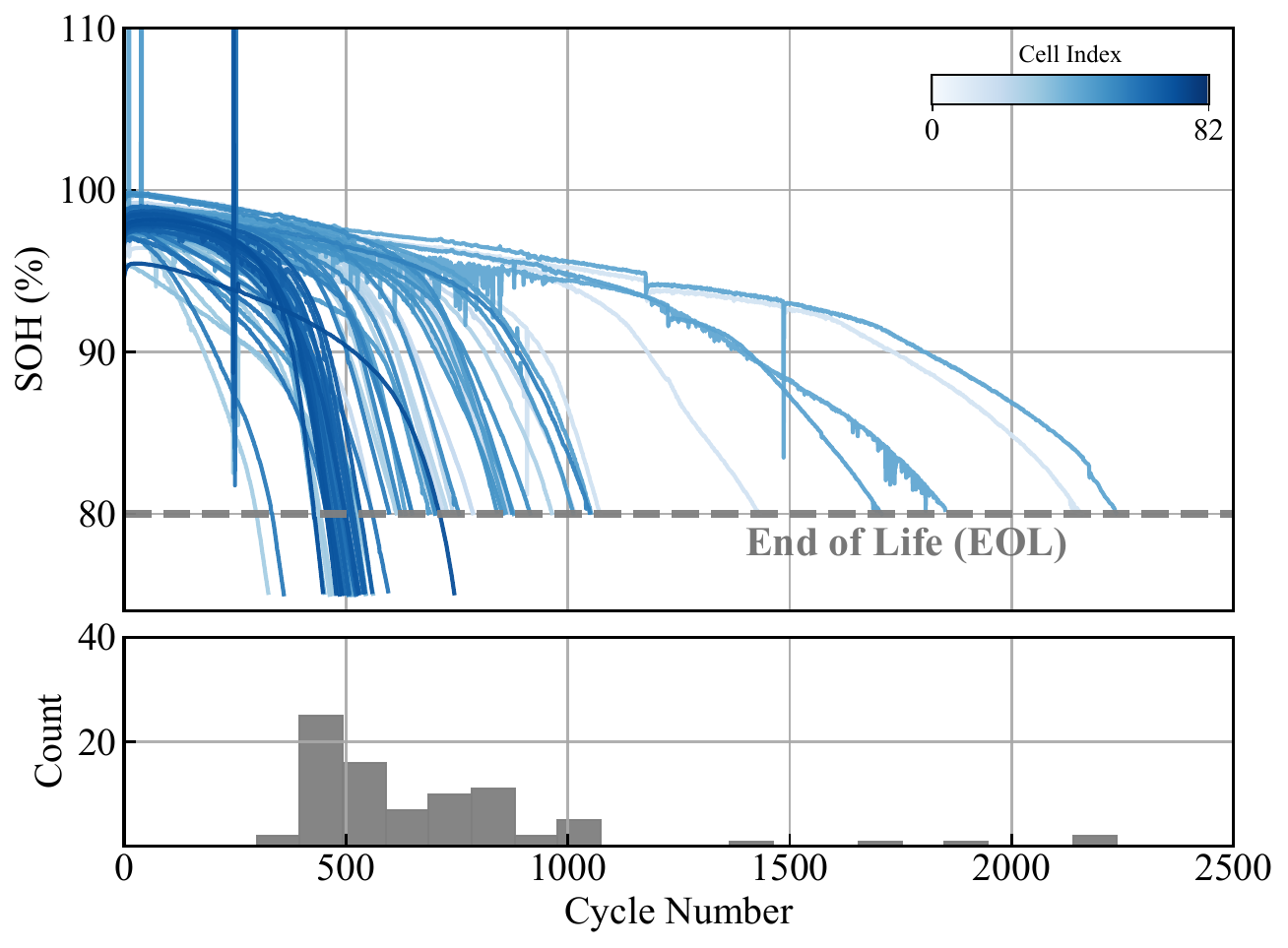}
        \caption{MATR1}
        \label{fig:dataset_matr1}
    \end{subfigure}
    \hfill
    \begin{subfigure}[t]{0.48\textwidth}
        \centering
        \includegraphics[width=\linewidth]{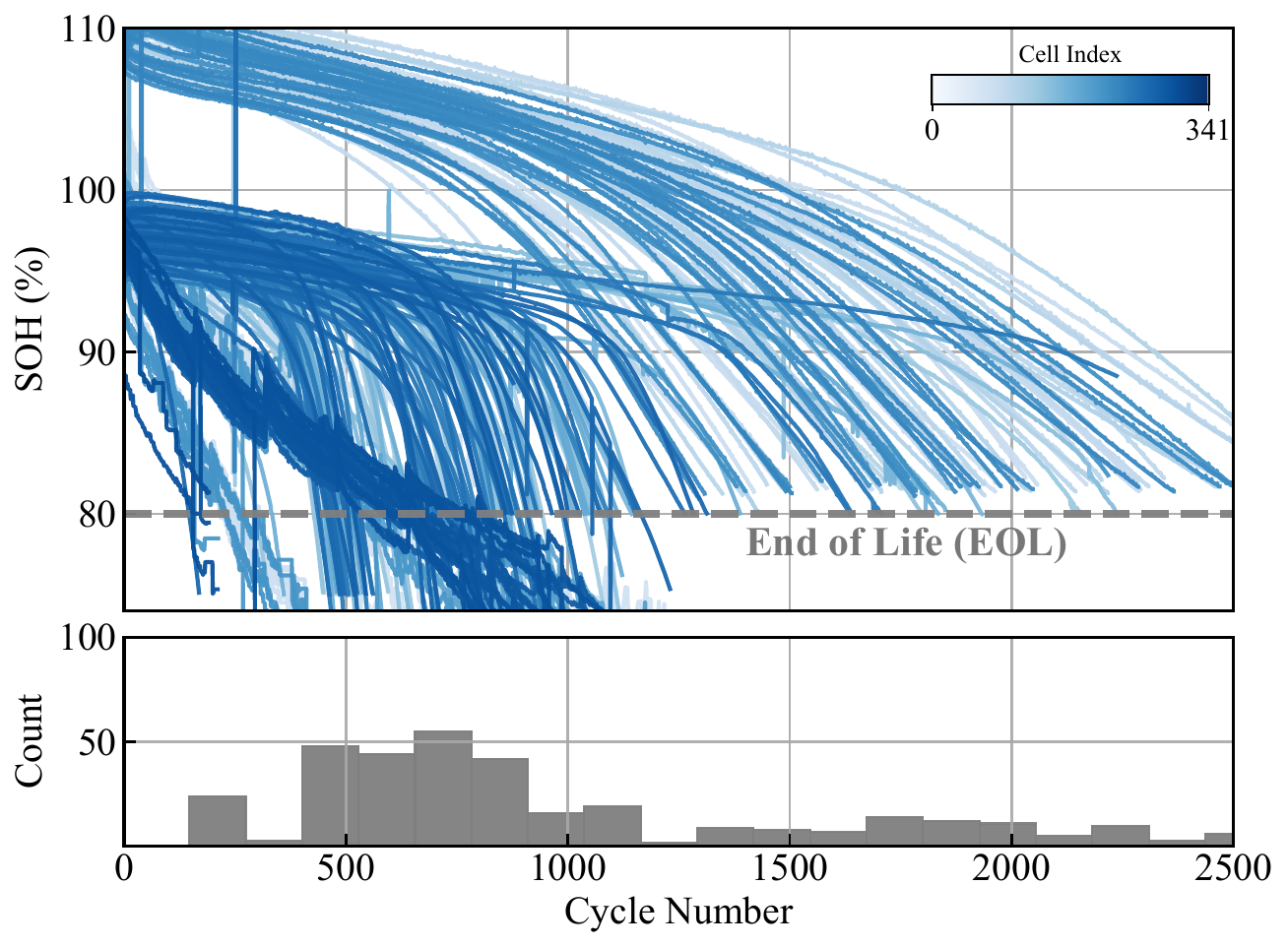}
        \caption{MIX}
        \label{fig:dataset_mix}
    \end{subfigure}
    
    \caption{Degradation trajectories and EOL distributions of cells in (a) MATR1 and (b) MIX benchmarks.}
    \label{fig:dataset}
\end{figure}
}

\newcommand{\ExpDatasetTab}{
\begin{table*}[\textwidth]
\centering
\caption{Characteristics of the battery datasets used for RUL prediction. The subscripts denote the standard deviation.}
\label{tab:dataset}
\small
\setlength{\tabcolsep}{4.8pt}
\renewcommand{\arraystretch}{1.2}
\begin{tabular}{lccccccc}
\toprule
Characteristics & CALCE & HNEI & MATR & UL\_PUR & SNL & RWTH & HUST \\
\midrule
Anode material & graphite & graphite & graphite & graphite & graphite & graphite & graphite \\
Cathode material & LCO & NMC/LCO & LFP & NCA & NCA/NMC/LFP & NMC & LFP \\
Form factor & Prismatic & Cylindrical  & Cylindrical  & Cylindrical  & Cylindrical  & Cylindrical  & Cylindrical  \\

Voltage range (V) & 2.7-4.2 & 3.0-4.3 & 2.0-3.5 & 2.7-4.2 & 2.0-4.2 & 3.5-3.9 & 2.0-3.6 \\
Nominal capacity (Ah) & $1.23_{0.12}$ & $2.80_{0}$ & $1.10_{0}$ & $3.40_{0}$ & $2.39_{0.88}$ & $1.11_{0}$ & $1.1_{0}$ \\
RUL distribution (cycles)& $566_{106}$ & $248_{15}$ & $823_{368}$ & $209_{50}$ & $1256_{1321}$ & $658_{64}$ & $1899_{389}$ \\

Cell count & 13 & 15 & 180 & 10 & 22/21/18 & 48 & 77 \\
\bottomrule
\end{tabular}
\end{table*}
}

\newcommand{\ExpRULLowCycleTab}{
\begin{table*}[t] 
\centering
\caption{Remaining useful life prediction results on benchmark datasets, evaluated at each method's minimum number of input cycles. “$>$1000” indicates that the RMSE exceeds 1000, reflecting large prediction errors. The subscripts denote the standard deviation.}
\label{tab:rul_low_cycle}
\small
\setlength{\tabcolsep}{0pt} 
\begin{tabular*}{\textwidth}{@{\extracolsep{\fill}} ll*{10}{c}}
    \toprule
    & Model & Cycle & MATR1 & MATR2 & HUST & SNL & CLO & CRUH & CRUSH & MIX & Avg \\
    \midrule
    & Ridge regression & 2 & 356.27 & 500.72 & $>$1000 & 417.92 & 515.33 & 325.29 & 601.98 & 576.19 & 576.24 \\
    & PCR              & 2 & 383.80 & 514.74 & 419.91 & 458.45 & 318.31 & 182.59 & 569.58 & 567.52 & 426.86 \\
    & PLSR             & 2 & 374.91 & 521.74 & 431.05 & 476.05 & 313.76 & 173.32 & 567.56 & 581.83 & 430.03 \\
    & Gaussian process & 2 & 356.27 & 500.73 & $>$1000 & 406.84 & 515.33 & 199.14 & $>$1000 & $>$1000 & $>$1000 \\
    & XGBoost          & 2 & 330.50 & 352.60 & 587.05 & 489.28 & 291.38 & 152.43 & 531.99 & 381.00 & 389.53 \\
    & Random forest    & 2 & 327.72 & 363.39 & 447.37 & 401.77 & 286.46 & 159.75 & 531.95 & 407.33 & 365.72 \\
    
    \cmidrule(lr){2-12}
    & CNN         & 2 & 358.73$_{6}$ & 463.80$_{3}$ & 398.99$_{48}$ & 465.05$_{257}$ & 298.12$_{369}$ & 155.92$_{134}$ & 576.22$_{2}$ & 359.67$_{109}$ & 384.56$_{116}$ \\
    & LSTM        & 2 & 302.66$_{8}$ & 360.98$_{47}$ & 419.95$_{41}$ & 386.17$_{65}$ & 304.59$_{14}$ & 173.77$_{14}$ & 567.98$_{25}$ & 474.54$_{30}$ & 373.83$_{31}$ \\
    & MLP         & 2 & 344.10$_{3}$ & 442.52$_{31}$ & 419.63$_{15}$ & 415.82$_{69}$ & 300.16$_{8}$ & 162.39$_{5}$ & 573.82$_{16}$ & 510.59$_{34}$ & 396.13$_{23}$ \\
    & Transformer & 2 & 260.24$_{10}$ & 422.55$_{28}$ & 572.32$_{11}$ & 402.17$_{17}$ & 269.70$_{7}$ & 130.51$_{15}$ & 504.53$_{13}$ & 381.48$_{25}$ & 367.94$_{16}$ \\
    
    \midrule
    & BatLiNet      & 2 & 139.52$_{8}$ & 397.01$_{14}$ & 351.01$_{39}$ & 282.08$_{16}$ & 219.04$_{11}$ & 131.63$_{5}$ & 457.08$_{17}$ & 276.95$_{58}$ & 281.79$_{21}$ \\
    & {CPMLP}        & {1} & {232.69$_{30}$} & {410.58$_{9}$} & {368.09$_{9}$} & {278.79$_{10}$} & {130.61$_{4}$} & {90.74$_{2}$} & {434.92$_{34}$} & {181.59$_{10}$} & {277.76$_{14}$} \\
    & {CPTransformer} & {1} & {232.34$_{29}$} & {448.84$_{29}$} & {369.32$_{3}$} & {299.39$_{19}$} & {123.04$_{7}$} & {91.50$_{2}$} & {392.18$_{52}$} & {188.41$_{10}$} & {268.13$_{19}$} \\
    \midrule
    & Ours & 1 & \textbf{101.69}$_{7}$ & \textbf{290.58}$_{12}$ & \textbf{345.31}$_{14}$ & \textbf{275.10}$_{44}$ & \textbf{102.63}$_{3}$ & \textbf{90.44}$_{3}$ & \textbf{359.45}$_{17}$ & \textbf{179.68}$_{8}$ & \textbf{218.11}$_{13}$ \\
    \bottomrule
\end{tabular*}
\end{table*}
}

\newcommand{\ExpavgTab}{
\begin{table}[width=\columnwidth]
\centering
\caption{Comparison of average RMSE, MAE, and MAPE across all benchmark datasets, using the fewest input cycles each method requires.}
\label{tab:avg_eval_table}
\begin{tabular}{@{}lcccc@{}}
\toprule
model            & Cycle & RMSE               & MAE                & MAPE               \\ \midrule
Ridge regression & 2     & 576.24             & 371.40             & 0.51               \\
PCR              & 2     & 426.86             & 306.49             & 0.54               \\
PLSR             & 2     & 430.03             & 298.71             & 0.48               \\
Gaussian process & 2     & \textgreater{}1000 & \textgreater{}1000 & \textgreater{}1.00 \\
XGBoost          & 2     & 389.53             & 254.24             & 0.38               \\
Random forest    & 2     & 365.72             & 232.90             & 0.29               \\ \midrule
CNN              & 2     & 384.56             & 270.71             & 0.49               \\
LSTM             & 2     & 373.83             & 250.55             & 0.39               \\
MLP              & 2     & 396.13             & 272.60             & 0.48               \\
Transformer      & 2     & 367.94             & 255.98             & 0.37               \\ \midrule
BatLiNet         & 2     & 281.79             & 188.89            & 0.28               \\ 
 CPMLP &  1 &  277.76 &  182.22 &  0.22 \\
 CPTransformer &  1 &  268.13 &  182.10 &  0.28 \\
\midrule
Ours             & 1     & \textbf{218.11}    & \textbf{144.02}    & \textbf{0.20}      \\ \bottomrule
\end{tabular}
\end{table}
}

\newcommand{\ExpTrainingTime}{
\begin{table}[!h]
\centering
\caption{Comparison of model complexity and training cost with and without scale compression under the 1-cycle and 100-cycle settings. Parameters (M) denotes the number of trainable model parameters in millions, and GFLOPs are reported on a per-sample basis.}
\label{tab:training_time2}
\resizebox{\columnwidth}{!}{
\begin{tabular}{@{}cc >{}c >{}c c@{}}
\toprule
Compression & Cycle & Parameters (M) & GFLOPs & Training time (s) \\
\midrule
             & 1    & 11.47 & 0.40  & 313.48  \\
\checkmark   & 1    & 11.32 & 0.36  & 225.84  \\ \midrule
             & 100  & 13.73 & 38.43 & 1522.05 \\
\checkmark   & 100  & 11.32 & 4.56  & 240.06  \\ \bottomrule
\end{tabular}
}
\end{table}
}

\newcommand{\ExpHeatmapMetricsFig}{
\begin{figure}[t]
    \centering
    \begin{subfigure}{\columnwidth}
        \centering
        \caption{RMSE}\label{fig:hm_rmse}
        \includegraphics[width=\linewidth]{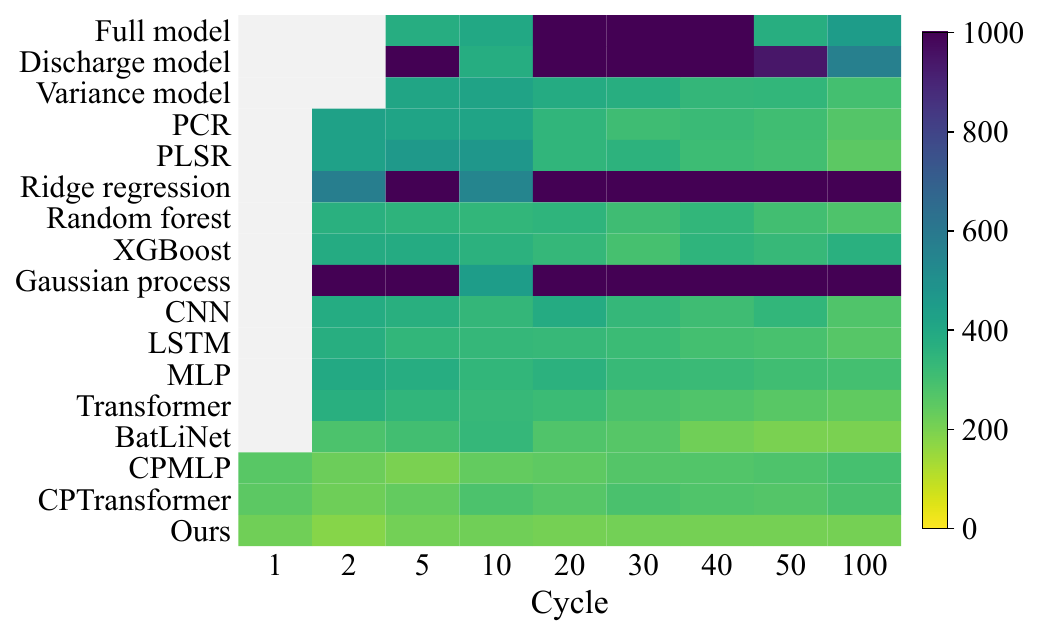}
    \end{subfigure}

    \vspace{2pt}

    \begin{subfigure}{\columnwidth}
        \centering
        \caption{MAE}\label{fig:hm_mae}
        \includegraphics[width=\linewidth]{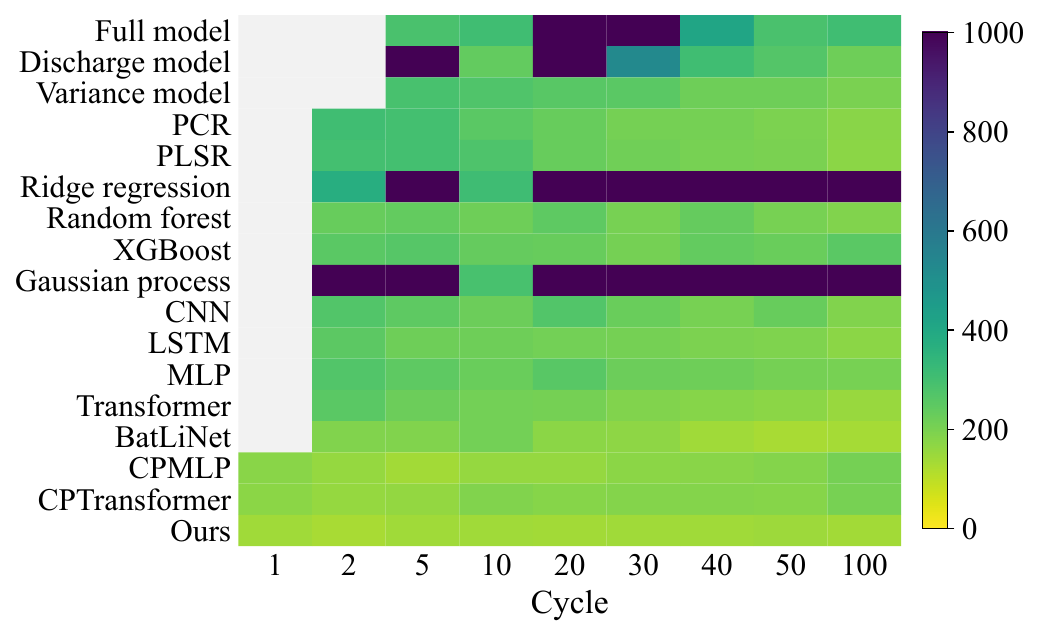}
    \end{subfigure}

    \vspace{2pt}

    \begin{subfigure}{\columnwidth}
        \centering
        \caption{MAPE}\label{fig:hm_mape}
        \includegraphics[width=\linewidth]{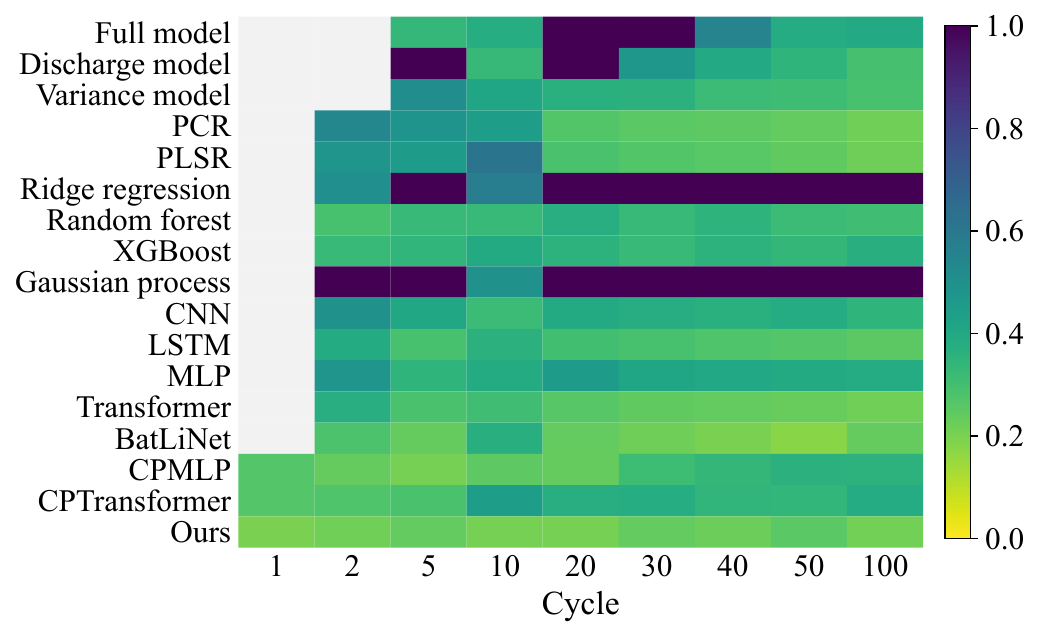}
    \end{subfigure}

    \caption{Heatmap showing average RMSE, MAE, and MAPE across benchmarks as the number of input cycles increases. White cells indicate missing settings (e.g., methods requiring $\geq2$ or $\geq5$ cycles). Our method demonstrates robustness across varying numbers of input cycles.}
    \label{fig:heatmap_metrics}
\end{figure}
}

\newcommand{\ExpwwoCWTFig}{
\begin{figure}[!t]
    \centering
    \includegraphics[width=0.95\linewidth]{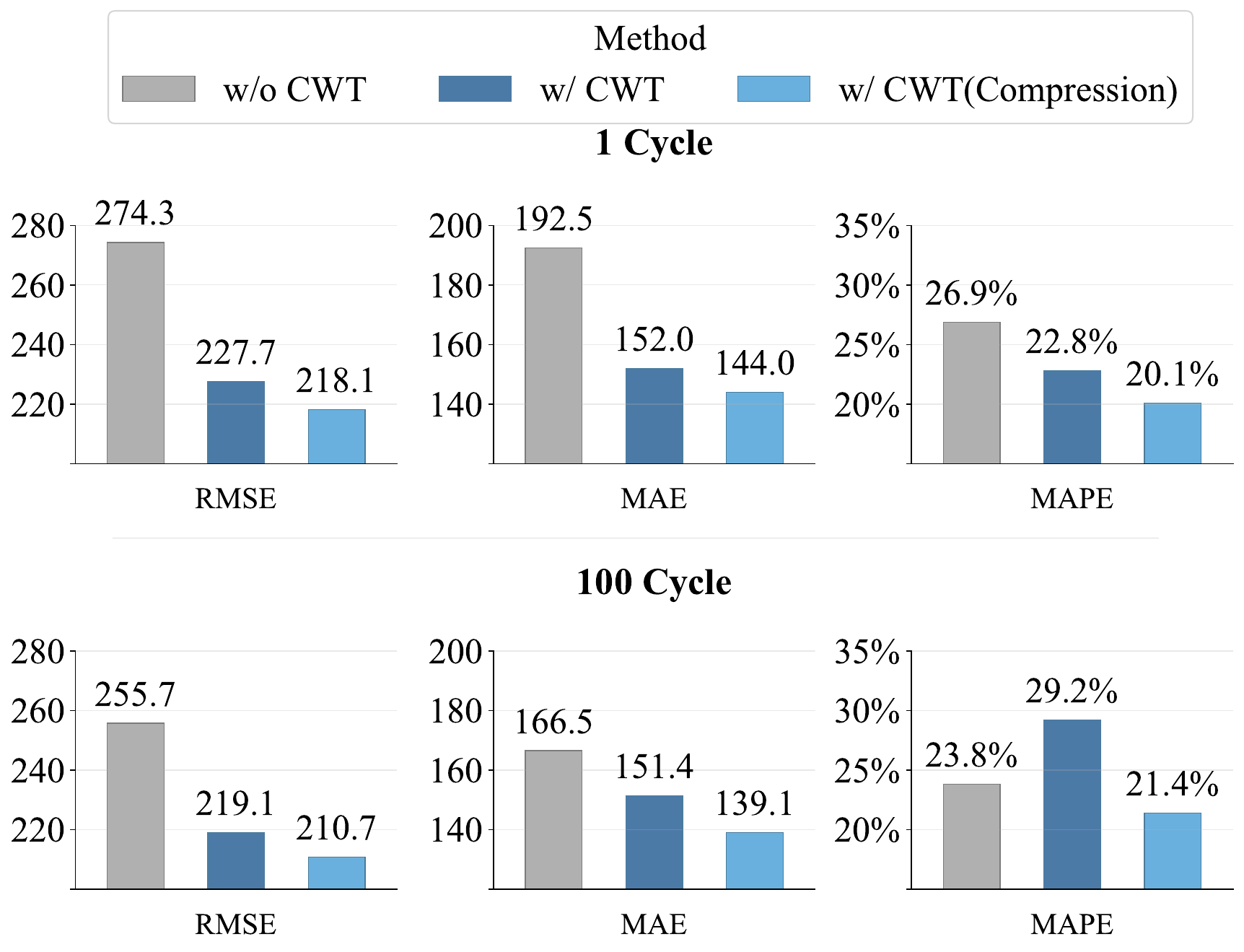}
    \caption{Comparison of average prediction error across three input configurations: time-domain, CWT, and CWT with scale compression.
    }
    \label{fig:wwoCWT}
\end{figure}
}

\newcommand{\ExpCompressMethodTab}{
\begin{table}[h!]
\centering
\caption{Ablation results for different scale compression methods. }
\label{tab:compression_method}
\begin{tabular}{@{}ccccc@{}}
\toprule
Method                           & Cycle & RMSE   & MAE    & MAPE \\ \midrule
\multirow{2}{*}{PCA}             & 1     & 331.33 & 233.77 & 0.43 \\
                                 & 100   & 266.65 & 183.25 & 0.33 \\ \midrule
\multirow{2}{*}{Top-$k$ energy}    & 1     & 309.75 & 218.77 & 0.48 \\
                                 & 100   & 294.99 & 213.15 & 0.38 \\ \midrule
\multirow{2}{*}{Energy-weighted} & 1     & 249.28 & 167.14 & 0.24 \\
                                 & 100   & 233.43 & 159.37 & 0.32 \\ \midrule
\multirow{2}{*}{Group-max}        & 1     & 246.89 & 166.69 & 0.24 \\
                                 & 100   & 226.27 & 153.66 & 0.24 \\ \midrule
\multirow{2}{*}{Group-mean}       & 1     & 218.11 & 144.02 & 0.20 \\
                                 & 100   & 210.72 & 139.07 & 0.21 \\ \bottomrule
\end{tabular}

\end{table}
}

\newcommand{\Exppredictplots}{
\begin{figure}[t]
    \centering
    \begin{subfigure}[t]{0.49\columnwidth}
        \includegraphics[width=\linewidth]{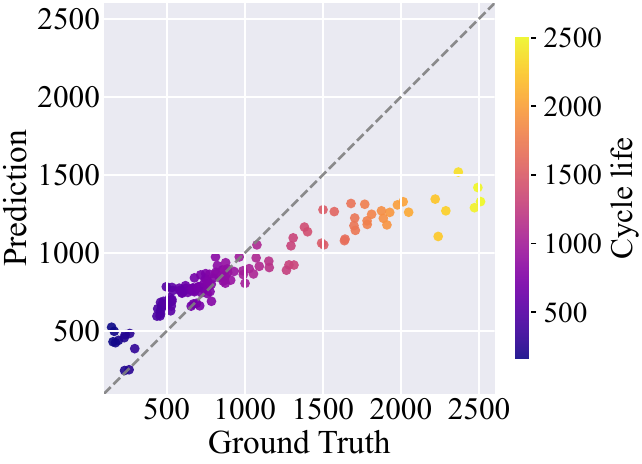}
        \caption{BatLiNet (2 cycle)}
        \label{subfig:batlinet_2}
    \end{subfigure}
    \hfill
    \begin{subfigure}[t]{0.49\columnwidth}
        \includegraphics[width=\linewidth]{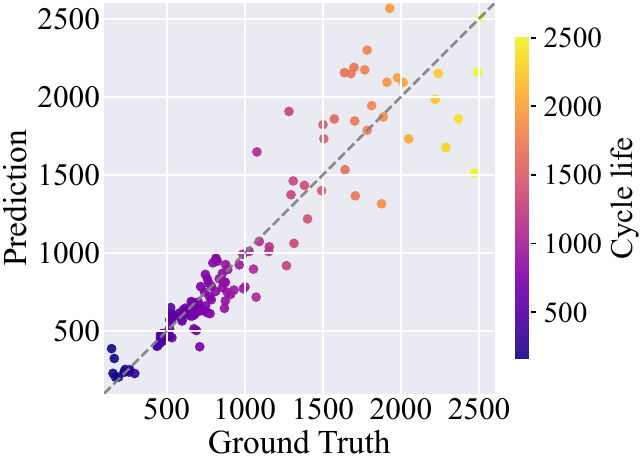}
        \caption{Ours (1 cycle)}
        \label{subfig:ours_1}
    \end{subfigure}
    
    \vspace{0.6em}
    
    \begin{subfigure}[t]{0.49\columnwidth}
        \includegraphics[width=\linewidth]{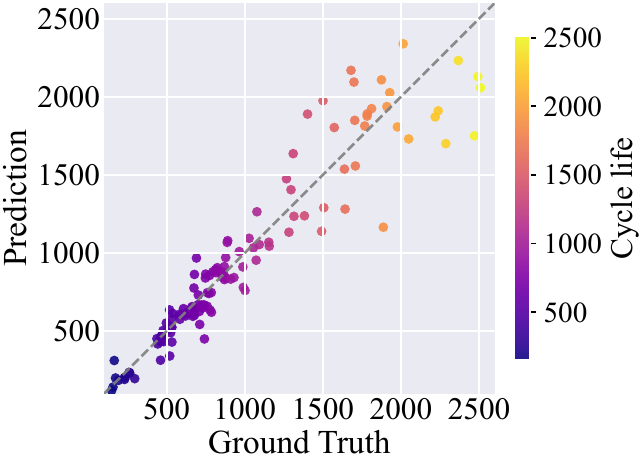}
        \caption{BatLiNet (100 cycles)}
        \label{subfig:batlinet_100}
    \end{subfigure}
    \hfill
    \begin{subfigure}[t]{0.49\columnwidth}
        \includegraphics[width=\linewidth]{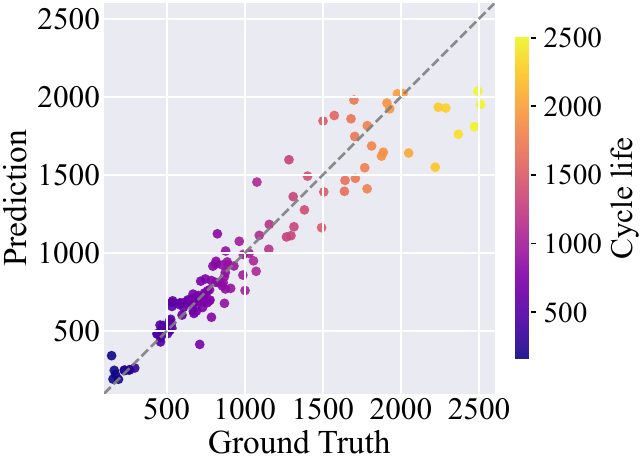}
        \caption{Ours (100 cycles)}
        \label{subfig:ours_100}
    \end{subfigure}
    
    \caption{Prediction results on the MIX dataset for BatLiNet and our method at their minimum input (top row; 2 and 1 cycles, respectively) and at 100 cycles (bottom row).}
    \label{fig:mix_prediction_all}
\end{figure}
}

\newcommand{\Expredundancy}{
\begin{figure}[t]
    \centering
    \begin{subfigure}{0.40\textwidth}
        \centering
        \includegraphics[width=\linewidth]{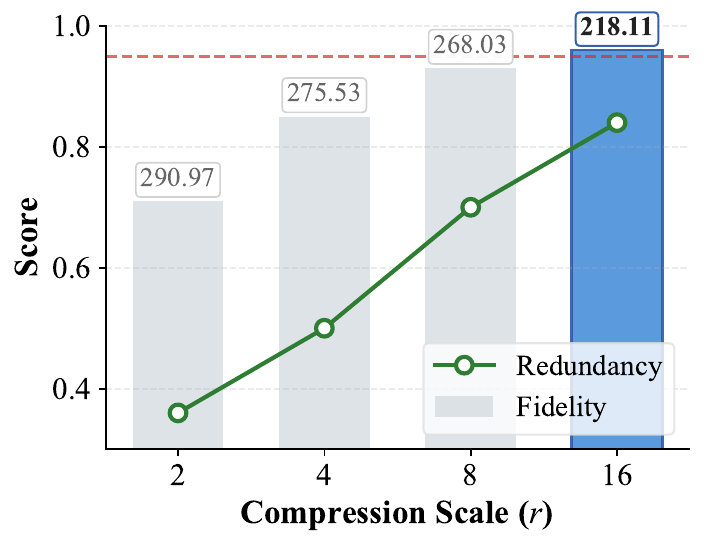}
        \caption{32-scale scalogram compression $r=2,4,8,16$.}
        \label{fig:compression_small}
    \end{subfigure}
    \hfill
    \begin{subfigure}{0.40\textwidth}
        \centering
        \includegraphics[width=\linewidth]{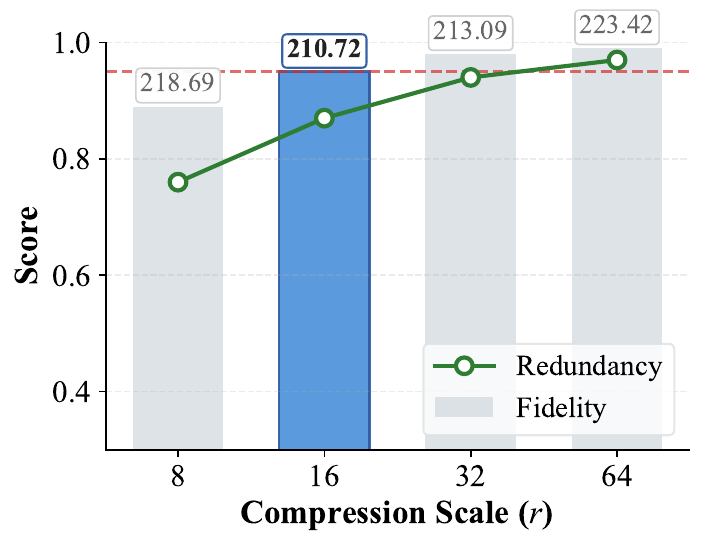}
        \caption{256-scale scalogram compression $r=8,16,32,64$.}
        \label{fig:compression_large}
    \end{subfigure}
    \caption{Trade-off between redundancy ratio and fidelity under different scale compression levels.}
    \label{fig:redundancy_fidelity}
\end{figure}
}

\newcommand{\Expablation}{
\begin{figure}[!h]
    \centering
    \includegraphics[width=\columnwidth]{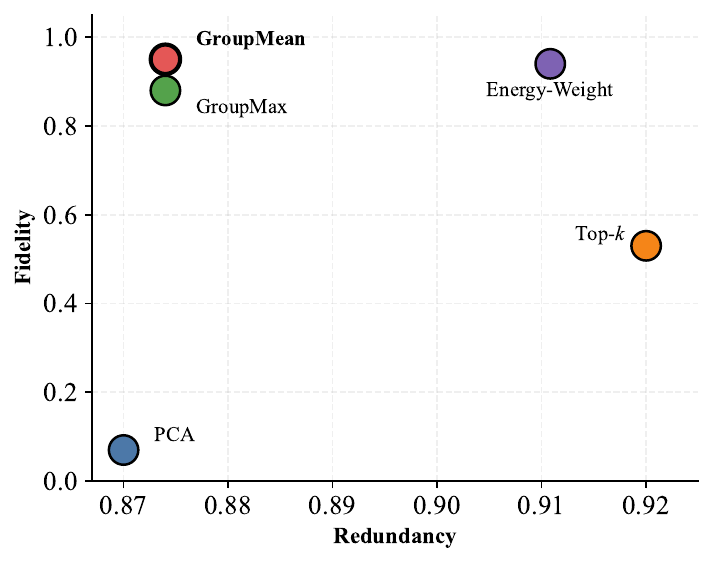}
    \caption{Trade-off between redundancy and fidelity across different scale compression strategies.
    }
    \label{fig:ablation_fid}
\end{figure}
}

\newcommand{\Expgradcam}{
\begin{figure*}[!ht]
    \centering

    \begin{subfigure}[t]{0.32\textwidth}
        \centering
        \includegraphics[width=\linewidth,clip]{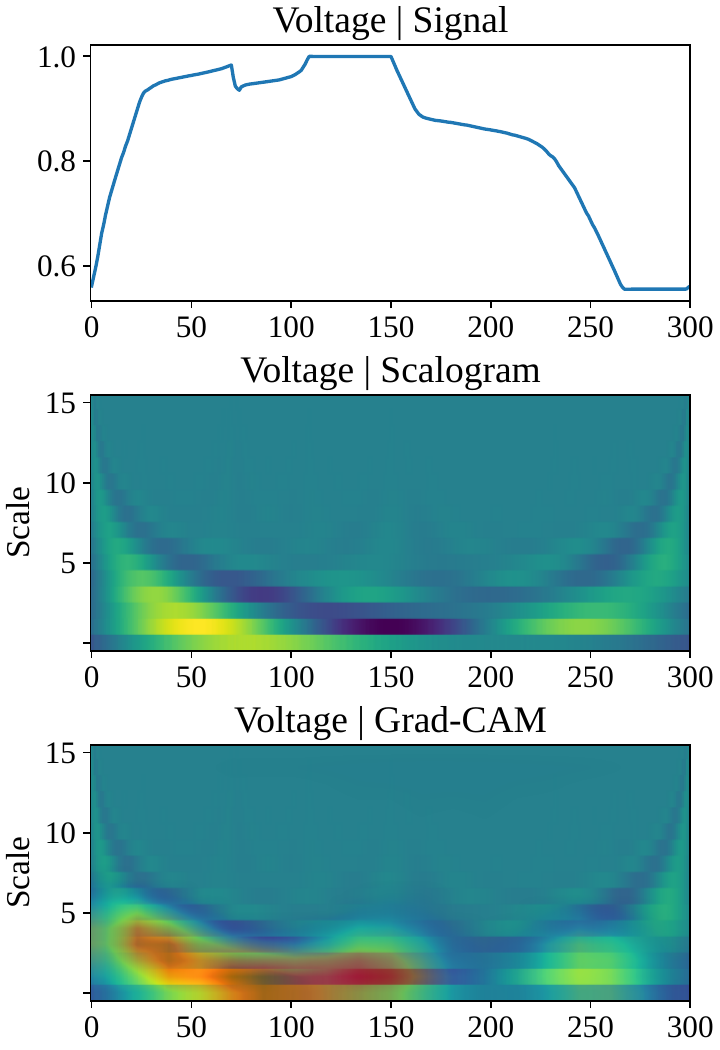}
        \caption{Voltage}
        \label{fig:gradcam_voltage}
    \end{subfigure}
    \hfill
    \begin{subfigure}[t]{0.32\textwidth}
        \centering
        \includegraphics[width=\linewidth,clip]{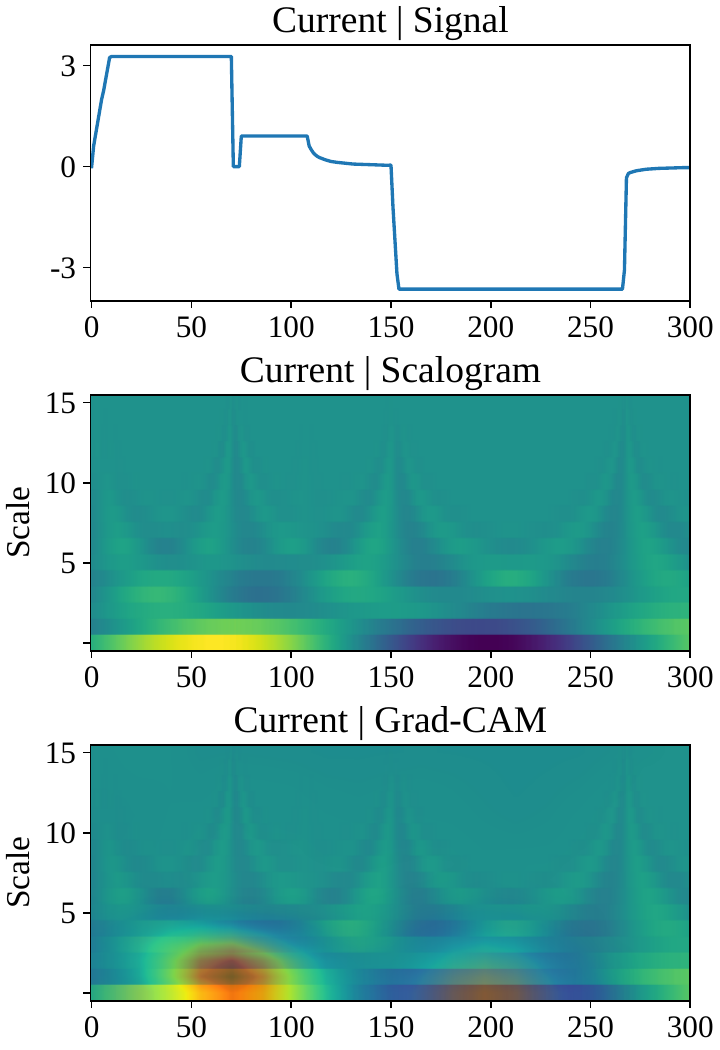}
        \caption{Current}
        \label{fig:gradcam_current}
    \end{subfigure}
    \hfill
    \begin{subfigure}[t]{0.32\textwidth}
        \centering
        \includegraphics[width=\linewidth,clip]{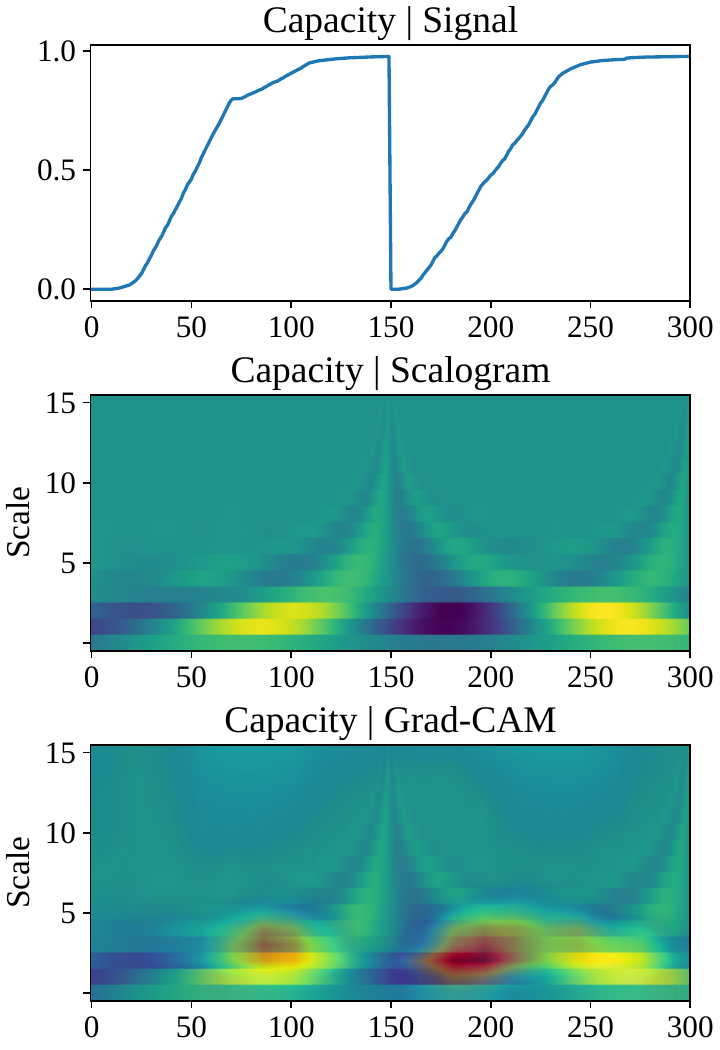}
        \caption{Capacity}
        \label{fig:gradcam_capacity}
    \end{subfigure}

    \caption{Grad-CAM visualization results for different input channels.
    (a) Voltage, (b) Current, and (c) Capacity. 
    Each panel shows the original signal, the corresponding CWT scalogram, and the Grad-CAM attention map.}
    
    \label{fig:gradcam_three_channels}
\end{figure*}
}
\newcommand{\Expviqdepend}{
\begin{table}[t]
\centering

\caption{RUL prediction performance on the MIX dataset under different input signal configurations. V, I, and C denote voltage, current, and capacity, respectively.}
\label{tab:viq_dependency}
\begin{tabular}{lccc}
\toprule
Input signals & RMSE & MAE & MAPE \\
\midrule
V   & 184.07 & 126.88 & 0.15 \\
I   & 182.30 & 124.93 & 0.14 \\
C   & 199.36 & 138.98 & 0.19 \\
\midrule
IC  & 186.10 & 129.61 & 0.15 \\
VC  & 191.34 & 145.80 & 0.19 \\
VI  & 186.36 & 130.24 & 0.15 \\
\midrule
VIC & \textbf{179.68} & \textbf{112.71} & \textbf{0.13} \\
\bottomrule
\end{tabular}
\end{table}
}

\newcommand{\Exptimefrequency}{
\begin{table}[h]
\centering

\caption{Comparison of time-frequency feature extraction methods for battery RUL prediction under the 1-cycle setting.}
\label{tab:timefrequency}
\begin{tabular}{lccc}
\toprule
Method & RMSE & MAE & MAPE \\
\midrule
Time-domain signal & 274.3 & 192.5 & 0.27 \\
STFT & 251.4 & 174.2 & 0.29 \\
DWT & 254.2 & 177.6 & 0.25 \\
\textbf{CWT} & \textbf{227.7} & \textbf{152.0} & \textbf{0.23} \\
\bottomrule
\end{tabular}
\end{table}
}

\newcommand{\Expkneepoint}{
\begin{figure}[t]
\centering

\begin{subfigure}{0.9\columnwidth}
\centering
\includegraphics[width=\linewidth]{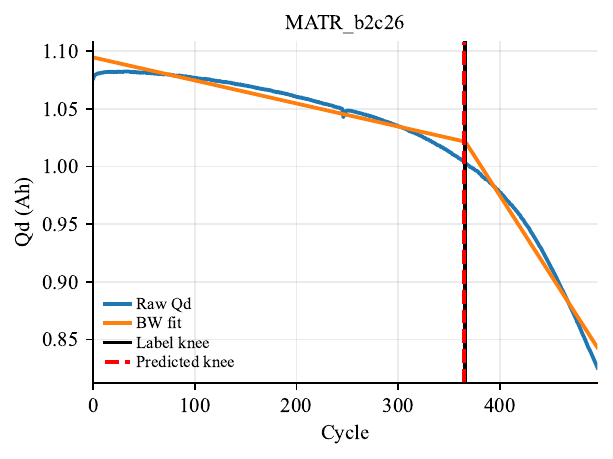}
\caption{Smooth degradation trajectory.}
\label{fig:knee_a}
\end{subfigure}

\begin{subfigure}{0.9\columnwidth}
\centering
\includegraphics[width=\linewidth]{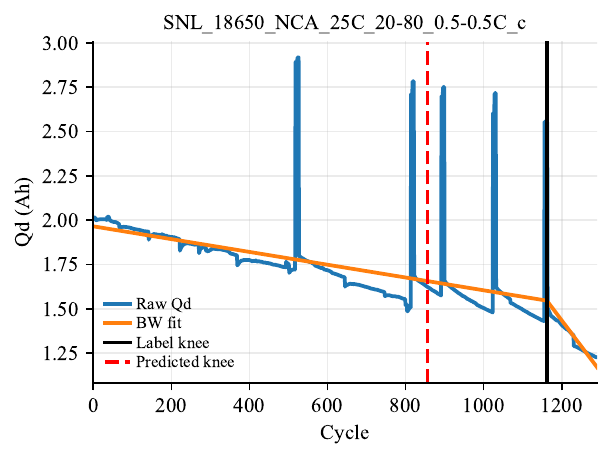}
\caption{Irregular spikes and ambiguous knee labeling.}
\label{fig:knee_b}
\end{subfigure}

\begin{subfigure}{0.85\columnwidth}
\centering
\includegraphics[width=\linewidth]{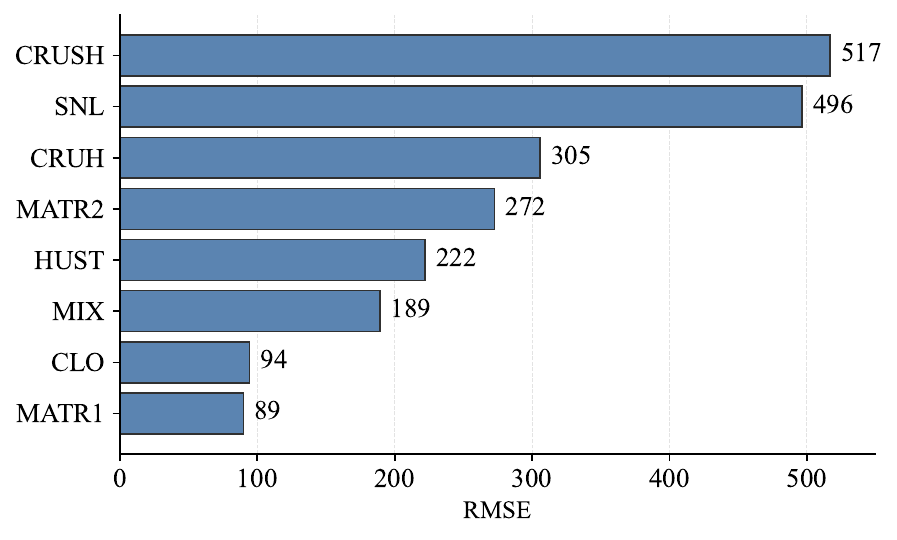}
\caption{Dataset-wise RMSE of knee-point prediction.}
\label{fig:knee_c}
\end{subfigure}

\caption{Auxiliary analysis of knee-point prediction under the single-cycle setting.}
\label{fig:knee_prediction_examples}

\end{figure}
}

\def\tsc#1{\csdef{#1}{\textsc{\lowercase{#1}}\xspace}}
\tsc{WGM}
\tsc{QE}
\tsc{EP}
\tsc{PMS}
\tsc{BEC}
\tsc{DE}

\begin{document}
\let\WriteBookmarks\relax
\def\floatpagepagefraction{1}
\def\textpagefraction{.001}
\shorttitle{Applied Energy}
\shortauthors{Sungeun Kim et~al.}

\title [mode = title]{Single-Cycle Battery Remaining Useful Life Prediction with Compressed Continuous Wavelet Features and Deep Neural Networks}

\author[1]{Sungeun Kim}[orcid=0009-0006-2819-8006]
\author[1]{Mingyu Jung}[orcid=0009-0001-7845-7284]
\author[1]{Yuri Kim}[orcid=0009-0007-0412-6224]
\author[2]{Jin-Sung Park}[orcid=0000-0001-5337-3722]
\author[1,3]{Jongbin Ryu}[orcid=0000-0001-5574-5358]
\cormark[1]
\affiliation[1]{organization={Department of Artificial Intelligence},
                addressline={Ajou University}, 
                city={Suwon},
                postcode={16499},
                country={Republic of Korea}}

\affiliation[2]{organization={Department of Materials Science \& Engineering},
                addressline={Ajou University}, 
                city={Suwon},
                postcode={16499},
                country={Republic of Korea}}

\affiliation[3]{organization={Department of Software \& Computer Engineering},
                addressline={Ajou University}, 
                city={Suwon},
                postcode={16499},
                country={Republic of Korea}}

\cortext[cor1]{Corresponding author}

\nonumnote{
\textcopyright{} 2026. This manuscript version is made
available under the CC BY-NC-ND 4.0 license
(\url{https://creativecommons.org/licenses/by-nc-nd/4.0/}).
The final published article is available at
\url{https://doi.org/10.1016/j.apenergy.2026.128798}.
}

\begin{abstract}
Lithium-ion batteries are widely used in industrial applications, but their performance gradually degrades over time, making accurate lifetime prediction important for safety and reliability. However, reliable lifetime assessment generally requires extensive degradation testing, which is both costly and time-consuming. To reduce this burden, recent data-driven approaches have sought to predict the end-of-life cycle of batteries from only the initial few cycles. Nevertheless, identifying reliable degradation patterns from such limited early-cycle data remains challenging because early aging dynamics are often subtle and nonlinear.
To address this issue, we adopt the continuous wavelet transform to convert cycle-wise voltage, current, and capacity signals into time-frequency representations, enabling multi-resolution characterization of degradation behavior. Building on this representation, we apply scale compression to reduce redundancy while preserving structural information in the scalograms.
We evaluate the proposed framework on eight benchmark datasets derived from seven public battery datasets covering diverse electrode chemistries, cell form factors, and cycling protocols.
Experimental results show that the proposed framework consistently outperforms prior approaches in single-cycle remaining useful life prediction and remains robust as the number of input cycles increases from 1 to 100.
In addition, redundancy-fidelity analysis characterizes the trade-off introduced by scale compression, and gradient-based visual explanations indicate that the model focuses on localized time-frequency regions associated with structured signal transitions during charge-discharge operation. These results demonstrate the effectiveness and interpretability of the proposed framework for early-cycle battery remaining useful life prediction.

\end{abstract}

\maketitle

\section{Introduction}
Lithium-ion batteries have been widely used in various industrial fields, including energy storage systems, electric vehicles, and portable electronic devices, due to their high energy density and efficient recharging. 
However, lithium-ion batteries exhibit significant performance degradation during repeated charge-discharge cycling, which eventually shortens battery life. Furthermore, batteries nearing the end of their operational life may experience abrupt voltage drops or thermal runaway events, posing significant risks to the safety and reliability of connected devices.

Therefore, it is important for battery manufacturers to mitigate degradation to produce long-lasting and durable lithium-ion batteries.
Directly evaluating battery life is the most accurate approach, but it requires long-term cycling tests that can take from hours to months. Because such tests are time-consuming and costly, conducting them extensively during battery development is often impractical.
For this reason, data-driven approaches have emerged as a promising alternative to traditional experimental testing for predicting the remaining useful life (RUL) of batteries.
These methods aim to predict the end-of-life (EOL), defined as the point when the nominal capacity falls below 80\%, using data collected during early charge-discharge cycles.

Among data-driven approaches, recent studies on battery lifetime prediction have explored a range of early-stage signals and modeling strategies. One line of research constructs domain-specific features from inter-cycle variations. Representative studies such as Severson \etal~\cite{severson2019data} and Attia \etal~\cite{attia2020closed} showed that handcrafted features derived from capacity differences across cycles can provide informative degradation descriptors for early-cycle lifetime prediction. These approaches offer interpretable features and can be effective under controlled conditions, but they typically require multiple comparable early cycles and may be less suitable when only extremely limited cycle data are available. 
Other studies have explored alternative signals, architectures, and learning paradigms for early-stage battery prediction. For example, Zhang \etal~\cite{zhang2024early} combined electrochemical impedance spectroscopy based geometric health indicators with regression and ensemble learning, Lee \etal~\cite{lee2024simultaneous} used convolutional neural network (CNN)- and recurrent neural network (RNN)-based models that jointly learn intra- and inter-cycle features for degradation and knee-point prediction, and Yang \etal~\cite{yang2021machine} applied machine learning to charging-process data under diverse operating conditions. These studies demonstrate the potential of data-driven modeling with different signals and architectures, although their applicability may depend on specific measurement modalities, richer cycle histories, or dataset-specific operating conditions.
Recent work has further advanced early-stage degradation-trajectory prediction and transfer/cross-domain learning~\cite{liu2025physics, lu2024hftl}, uncertainty-aware and physics-informed modeling~\cite{oyewole2025uncertainty}, and related RUL prediction or fault diagnosis in broader prognostics applications~\cite{hu2024m2bist, hu2026smnet}. Despite this broad progress, predicting RUL from a very small number of standard charge-discharge cycles remains challenging.
This challenge is particularly important because many early prediction settings still rely on several tens of cycles or the first 100 cycles as input. However, under standard laboratory protocols—e.g., 1C charge/discharge with constant-current constant-voltage (CC-CV) charging and rest periods for voltage relaxation—a battery degradation test requires at least two hours to complete a single cycle. Consequently, collecting 100 cycles typically requires over two weeks, incurring significant time and resource costs. 
In contrast, reliable RUL prediction from only the first cycle would compress this multi-week testing loop to a single cycle, providing the practical motivation for the single-cycle setting. This setting is challenging because the available observation is extremely limited.

\Expoverall
Nevertheless, recent studies suggest that even very early battery signals may already contain lifetime-relevant information before explicit capacity degradation becomes visible. For example, early-cycle electrochemical features derived from voltage responses have been reported to correlate with battery lifetime across multiple chemistries~\cite{paulson2022feature}, and prior work has also demonstrated the feasibility of battery RUL prediction from a single cycle~\cite{hsu2022deep}. These findings support the premise that meaningful degradation-related structure may already be embedded in limited early-cycle signals.

Battery cycling data typically consists of time-series measurements of voltage, current, and capacity collected during charge-discharge operations. These signals indirectly reflect the influence of underlying degradation mechanisms, such as electrode damage, growth of the solid electrolyte interphase (SEI) layer, and increased internal resistance~\cite{edge2021lithium}. In the single-cycle setting, the first cycle provides the earliest observable charge-discharge response of a cell under a given cycling protocol. Although this response does not directly reveal long-term capacity fade, it may contain signal-level variations associated with the initial electrochemical state, polarization behavior, protocol-dependent operating response, and manufacturing-induced cell-to-cell variability. In this sense, the first-cycle response can be regarded as a signal-level response in which these electrochemical, protocol-dependent, and manufacturing-related variations are jointly reflected, rather than as a direct identification of specific electrochemical phenomena or degradation mechanisms.
Because the initial cell state and protocol-dependent response can influence subsequent aging behavior, this response may provide informative predictive cues relevant to long-term battery lifetime. 
Furthermore, battery signals are inherently non-stationary, and degradation-related variations are often weak in the early stage of cycling. Therefore, extracting patterns from these limited observations remains challenging in the time domain alone.

In this context, time-frequency domain analysis of battery signals has been actively investigated as a promising alternative to conventional time-domain approaches~\cite{chae2025state, el2020time, zhang2022feature}.
Among the most widely used approaches for time-frequency analysis are the short-time Fourier transform (STFT) and wavelet transform. These techniques enable the simultaneous examination of time-frequency characteristics of a signal, making them effective and useful for fault diagnosis and prognostics, including RUL prediction.
Wavelet-based analysis has also been investigated in prognostic and diagnostic applications under challenging signal conditions~\cite{ibrahim2016wavelet,zheng2002gear}. These studies provide general motivation for considering wavelet-based time-frequency representations, although their applicability depends on the characteristics of the specific prediction task.
For instance, the wavelet transform has been applied to one-dimensional vibration signals for bearing RUL prediction, converting them into two-dimensional time-frequency representations that are effectively used with CNN. This approach achieves the lowest prediction error among the comparative studies, demonstrating its effectiveness in the RUL prediction~\cite{yoo2018novel}.
Similarly, wavelet-based time-frequency scalograms have been successfully utilized in deep learning models for epilepsy classification using the electroencephalogram signals, achieving high accuracy across various diagnostic tasks~\cite{turk2019epilepsy}. 

In the battery domain, wavelet-based time-frequency images derived from voltage signals have been employed with pre-trained CNN architectures through transfer learning, demonstrating improved accuracy and robustness in state-of-health (SOH) prediction~\cite{el2020time}.

Therefore, to capture battery degradation patterns using only a small number of initial cycles, we apply the multi-resolution continuous wavelet transform (CWT) to interpolated raw cycling signals (voltage, current, and capacity) for each cycle, thereby obtaining cycle-wise time-frequency representations.
This representation is particularly suitable for the single-cycle setting, where degradation cues are too weak to be reliably characterized by inter-cycle differences or direct time-domain modeling alone. Compared with purely time-domain features, the CWT can represent localized high-frequency variations and broader low-frequency patterns within the single-cycle response, making it useful for early-cycle RUL prediction.

The CWT requires more computation time than the discrete wavelet transform (DWT) because of its higher computational complexity. However, it represents the transformed signal as a scalogram, which organizes the single-cycle response across time and scale. This representation enables CNN-based feature extraction of localized patterns within the scalogram. In this study, the CWT is used to represent weak and non-stationary signal-level variations within a single charge-discharge response, rather than to directly decompose specific degradation mechanisms.
At the same time, the densely sampled scale axis of the CWT can produce large representations, and adjacent scale ranges often contain highly similar information. This increases computational overhead during model training and introduces substantial internal redundancy. To address this issue, we apply scale compression that aggregates adjacent scale coefficients into representative values.
The overall framework, illustrating this compression process within our RUL prediction pipeline, is shown in~\Cref{fig:framework}.
The goal is not merely to reduce dimensionality, but to reduce redundancy while preserving as much as possible the informative structure of the original time-frequency representation. We further analyze this trade-off quantitatively in terms of redundancy reduction and fidelity preservation in \cref{subsec:redundancy}. 
Taken together, these considerations motivate a CWT-based pipeline with scale compression as an integrated representation strategy for the single-cycle setting, where inter-cycle differences are unavailable and observable lifetime-related cues are limited.

We evaluate the proposed method on eight benchmark datasets derived from seven public battery datasets, encompassing diverse electrode chemistries, cell form factors, and cycling protocols, using ResNet~\cite{he2016deep}, a widely adopted CNN architecture. In the single-cycle setting, the proposed method achieves an average RMSE of 218.11 using only the first cycle as input. The results further show that CWT-based feature extraction improves prediction performance over direct time-domain input, and that scale compression provides additional gains in both prediction accuracy and computational efficiency compared with uncompressed CWT inputs. In addition, experiments with varying numbers of input cycles confirm the robustness of the proposed method from 1 to 100 cycles. 
We further examine the relative contributions of voltage, current, and capacity inputs and use explainable artificial intelligence (XAI) to interpret which regions of the time-frequency representation are most informative for prediction. 
These results indicate that the proposed pipeline can extract useful signal-level patterns for RUL prediction, even when only extremely limited early-cycle observations are available.

Our main contributions are as follows:
\begin{itemize}
    \item We develop a compact representation pipeline for single-cycle battery RUL prediction by converting cycle-wise battery signals into scale-compressed CWT scalograms for CNN-based learning.
    
    \item We incorporate scale compression into the CWT-based representation to reduce inter-scale redundancy and computational cost while preserving the informative structure of the scalograms.
    
    \item We comprehensively evaluate the proposed framework on eight benchmark datasets derived from seven public battery datasets, demonstrating strong single-cycle RUL prediction performance, robustness across varying input cycles, and interpretability through input-dependence and XAI-based analyses.
\end{itemize}

\section{Methodology}
\subsection{Compressed continuous wavelet features} 
Time-frequency analysis methods are widely employed to investigate the intrinsic characteristics of time-series signals by transforming them from the time domain to the frequency domain. While the Fourier transform effectively reveals global frequency components, it lacks temporal resolution, making it less suitable for analyzing non-stationary signals such as degradation signals from lithium-ion batteries, whose statistical properties evolve over time. As a result, various time-frequency techniques capable of capturing such non-stationary behaviors have been developed~\cite{yao2020novel, di2024battery, kim2023time}.
The wavelet transform decomposes a signal into the time-frequency domain, enabling precise capture of localized variations. Its capability to analyze structural features across multiple scales makes it a widely adopted and effective tool for detecting and interpreting non-stationary signals~\cite{torrence1998practical}.
Therefore, we apply the CWT to analyze the time-frequency characteristics of battery data at multiple resolutions. The CWT provides a time-frequency scalogram with better resolution than the DWT, which uses dyadic levels instead of continuous scales~\cite{bolos2014wavelet}. For an input signal $x(t)$, the CWT is defined as follows.

\begin{equation}
W_x(a, b) = \frac{1}{\sqrt{|a|}} \int_{-\infty}^{\infty} x(t)\, \psi\left( \frac{t - b}{a} \right) dt,
\end{equation}
where $\psi(t)$ denotes the mother wavelet function, $a \in \mathbb{R}^+$ denotes the scale parameter, and $b \in \mathbb{R}$ is the time shift parameter. The wavelet coefficient $W_x(a, b)$ is obtained by projecting the input signal $x(t)$ onto a wavelet that has been scaled by $a$ and shifted by $b$, and it captures the localized time-frequency content of the signal at that specific scale and time.
We employ the Morlet wavelet as the mother wavelet due to its favorable balance between time and frequency resolution, as well as its demonstrated effectiveness in analyzing non-stationary signals~\cite{cohen2019better}. The Morlet wavelet is defined as follows:
\begin{equation}
\psi(t) = \exp^{j \omega_0 t} \cdot \exp^{-t^2 / (2\sigma^2)},
\end{equation}
where $\omega_0$ denotes the center frequency of the wavelet, and $\sigma$ represents the standard deviation of the Gaussian window.
In the wavelet transform, the scale parameter controls the temporal extent of the wavelet function, enabling the extraction of multi-resolution information through signal analysis across different scales. We perform multi-resolution analysis on the input signal $x(t)\in\mathbb{R}^T$ using a scale set $\mathcal{A} = \{a_1<a_2<\cdots<a_m\}\subset \mathbb{R}^+$. Let $\mathcal{B} = \{b_1<b_2<\cdots<b_t\}\subset \mathbb{R}$ denote the set of time shift parameters. Then, the scalogram $S_x$ obtained via wavelet transform is defined as follows.
\begin{equation}
\label{eq:scalogram}
\mathbf{S}_x = \left\{ \left| W_x(a, b) \right| \;\middle|\; a \in \mathcal{A},\; b \in \mathcal{B} \right\}.
\end{equation}

The wavelet transform enables multi-resolution analysis of signals by adjusting the scale, thereby providing different time-frequency resolutions. The Morlet wavelet, a non-orthogonal function, is not restricted to a fixed set of discrete scales. Instead, it allows flexible selection of continuous scales, enabling precise resolution control and adaptable analysis tailored to specific data conditions~\cite{torrence1998practical}.
The characteristics of resolution vary depending on the scale, with smaller scales reflecting high-frequency components over short time spans and providing high temporal resolution. In comparison, larger scales reflect low-frequency components over longer durations and offer high frequency resolution~\cite{najmi1997continuous}.
This enables precise multi-resolution analysis of input signals. However, adjacent scales often contain redundant information, which can increase the presence of noise and other irrelevant features~\cite{peter2004machine}. 
Furthermore, using a scalogram generated over a wide range of scales as input to a deep learning model can significantly increase input dimensionality and computational cost.
To address this issue, we apply scale compression to reduce feature redundancy while minimizing information loss from the wavelet transform.
In this strategy, the full set of scales $\mathcal{A}$ is uniformly divided into $r$ adjacent groups $\{G_g\}^r_{g=1}$. Each group $G_g$ is defined by the range of scale indices $k$ as follows.
\begin{equation}
\label{eq:scalecompression}
G_g = \left\{ a_k \in \mathcal{A} \;\middle|\; \left\lceil \frac{(g - 1) \cdot m}{r} + 1 \right\rceil \leq k \leq \left\lfloor \frac{g \cdot m}{r} \right\rfloor \right\},
\end{equation}
here, $r$ denotes the number of compressed scales, with $r \leq m$, where $m$ is the total number of original scales.
Each group ${G_g}$ consists of adjacent wavelet coefficients, and the compressed coefficient $\tilde{W}_x(g, b)$ is computed as the group mean, defined as follows:
\begin{equation}
\label{eq:group}
\tilde{W}_x(g, b) = \frac{1}{|G_g|} \sum_{a \in G_g} W_x(a, b),
\end{equation}
since \(m\) is divisible by \(r\) in all evaluated scale configurations, the group-mean operation in~\cref{eq:group} is mathematically equivalent to non-overlapping one-dimensional average pooling along the scale axis, with both the kernel size and stride set to \(m/r\).
Accordingly, the final compressed scalogram $\tilde{\mathbf{S}}_x$ derived from the wavelet transform is given by:
\begin{equation}
\label{eq:compressed_scalogram}
\tilde{\mathbf{S}}_x = \left\{ \left| \tilde{W}_x(g, b) \right| \;\middle|\; g=1,2,\ldots,r,\; b \in \mathcal{B} \right\}.
\end{equation}

Accordingly, we apply the CWT to perform multi-resolution analysis of battery cycling signals and use scale compression based on group-wise averaging to mitigate the feature redundancy introduced by densely sampled time-frequency representations. This aggregation reduces redundancy across neighboring scales while minimizing information loss.
To determine the most effective compression strategy, we empirically evaluate five different methods and select the group mean-based approach based on its superior performance.
A detailed ablation study of these alternatives is provided in \cref{subsec:ablation}.

\subsection{Continuous wavelet feature extraction for single- and multi-cycle RUL prediction}
\label{subsec:wavelet_rul}
Battery capacity progressively degrades over time due to chemical and structural changes that occur with regular use.
To quantitatively evaluate this degradation under controlled conditions, batteries are subjected to degradation tests involving repeated charge-discharge cycles performed according to predefined protocols and regulated temperature environments.
The capacity measured in each cycle serves as the basis for computing the SOH, which is defined as:
\begin{equation}
    SOH = \frac{C_i}{C_0}
\end{equation}
where $C_0$ denotes the nominal capacity, and $C_i$ is the discharge capacity measured at the $i$-th cycle.
In general, the EOL for RUL prediction tasks is defined as the point at which SOH falls below 80\% of the nominal capacity.

Measuring battery data through degradation testing until the SOH reaches the EOL provides an accurate assessment of the battery lifespan. However, such testing is extremely time- and cost-intensive. As a result, recent studies have actively explored RUL prediction using limited data obtained from early cycles.
Severson \etal~\cite{severson2019data} observed that changes in discharge capacity during the early cycles of battery operation are minimal, which limits the effectiveness of using raw data directly for lifetime prediction. To address this, they proposed a domain-specific feature extraction method by defining an interpolated voltage-capacity function $Q_n(V)$ for the n-th cycle and calculating the differential curve $\Delta Q_{n-m}(V)$, which quantifies the capacity deviation between two cycles at a given voltage. In particular, the voltage-specific capacity difference between the 10th and 100th cycles, $\Delta Q_{100-10}(V)$, was utilized as an input feature, resulting in improved RUL prediction performance.

Building on this approach, Attia \etal~\cite{attia2021statistical} extended the methodology by using the full $\Delta Q_{100-10}(V)$ vector as input to various multivariate statistical models.
Furthermore, Zhang \etal~\cite{batteryml} incorporated $\Delta Q_{100-10}(V)$-based domain-specific features into deep learning models such as CNN and long short-term memory (LSTM), extending their use beyond statistical models and yielding competitive performance in nonlinear architectures for early-cycle RUL prediction.
However, since this approach relies on computing differences between multiple cycles, it requires comparable cycle pairs to construct the features. Consequently, its applicability may be limited when only a few early cycles are available, and predictive performance can degrade when sufficient comparable cycles are unavailable.

To address this issue, we adopt a cycle-wise feature extraction approach inspired by Tan \etal~\cite{tan2025batterylife}. In this approach, the voltage, current, and capacity data from each cycle are resampled, linearly interpolated, and normalized. This method extracts features independently for each cycle and then transforms them into the time-frequency domain using the CWT. Specifically, for each cycle $i$, the voltage, current, and capacity signals are linearly interpolated to a fixed length $T$, resulting in the following input matrix:
\begin{equation}
X_i = \begin{bmatrix} x_i^{(v)}(t) \\ x_i^{(j)}(t) \\ x_i^{(c)}(t) \end{bmatrix} \in \mathbb{R}^{3 \times T}, \quad i = 1, 2, \dots, N ,
\label{eq:raw_extract}
\end{equation}
where $x_i^{(v)}(t)$, $x_i^{(j)}(t)$, and $x_i^{(c)}(t)$ represent the interpolated voltage, current, and capacity signals of the $i$-th cycle, respectively. The sequential representation of multiple cycles from $i$ to $N$ is denoted as:
\begin{equation}
X_{i:N} = [X_i, X_{i+1}, \dots, X_N] \in \mathbb{R}^{3 \times (N - i + 1) \times T}.
\end{equation}

To extract the time-frequency characteristics of each cycle $X_i$, we apply the CWT individually to the interpolated voltage $x_i^{(v)}(t)$, current $x_i^{(j)}(t)$, and capacity $x_i^{(c)}(t)$ signals. The resulting scalograms are concatenated to form the input feature vector $F_i$. To reduce computational cost and feature redundancy, we apply the scale compression strategy described in \cref{eq:scalecompression}, in which the full scale set $\mathcal{A}$ is uniformly divided into $r$ adjacent groups, and the wavelet coefficients within each group are aggregated by averaging. Accordingly, the resulting feature representation $F_i$ is given by:

\begin{equation}
\begin{aligned}
F_i &=
\begin{bmatrix}
\tilde{S}_i^{(v)} \\
\tilde{S}_i^{(j)} \\
\tilde{S}_i^{(c)}
\end{bmatrix}
\in \mathbb{R}^{3r \times T},
\end{aligned}
\end{equation}
Each $\tilde{S}_i^{(h)}$, where $h \in \{v, j, c\}$, denotes the compressed scalogram corresponding to voltage, current, and capacity, respectively, and is computed according to \cref{eq:compressed_scalogram}. This representation allows the model to leverage information from a single cycle $i$ as input. Moreover, it can be extended to a sequence of consecutive cycles from $i$ to $N$ as follows:
\begin{equation}
F_{i:N} = [F_i, F_{i+1}, \dots, F_N] \in \mathbb{R}^{3r \times (N - i + 1) \times T}.
\end{equation}

The proposed representation enables the effective use of information from a single cycle or a specific range of cycles. This flexibility is particularly beneficial in early-cycle RUL prediction.
Moreover, our experiments reveal that tailoring the wavelet scale set to the number of available cycles significantly impacts the quality of the resulting time-frequency representations. 
When a sufficient number of early cycles are available, using a wide range of scales captures both low- and high-frequency components, yielding multi-resolution features. 
In contrast, for shorter inputs—such as a single cycle—the limited information content renders a broad scale range less effective, often introducing redundancy and degrading predictive performance.
To accommodate these different input lengths, we construct the wavelet scale set according to the exponential scale selection method of Torrence and Compo~\cite{torrence1998practical}, in which logarithmically spaced scales are determined from the input length and sampling interval. Under this rule, the scale range is restricted for short inputs to avoid excessive sensitivity to sub-cycle noise, while it is expanded for longer inputs to preserve broader multiscale information. Accordingly, we use 32 scales in the 1-cycle setting and 256 scales in the 100-cycle setting. These original scalograms are then compressed to a common scale dimension of $r=16$ using the scale compression method. A detailed quantitative analysis of the selected compression level, including the trade-off between redundancy reduction and fidelity preservation, is provided in \cref{subsec:redundancy}.

This scale adjustment strategy enables the practical application of the CWT even when the input consists of very short sequences, such as those from a single early cycle, thereby facilitating the extraction of meaningful time-frequency information from limited data. Furthermore, applying scale compression in conjunction with the exponential scale selection strategy, which is particularly effective for short input sequences such as single-cycle data, reduces feature redundancy and enhances RUL prediction performance. These effects are quantified in \cref{subsec:cwt}.

Overall, the proposed CWT-based representation with scale compression offers a practical feature extraction approach for RUL prediction using early-cycle data.
It effectively preserves the time-frequency characteristics of battery degradation signals while flexibly adapting to the limitations of short input sequences. Furthermore, when applied to a standard CNN architecture such as ResNet, it enables robust predictive performance without requiring any architectural modifications. These properties are evaluated across eight public benchmark datasets in the following section.

\section{Experiments}
\subsection{Experimental data}
\ExpDatasetTab
We use seven publicly available lithium-ion battery datasets in this study: CALCE~\cite{xing2013ensemble, he2011prognostics}, HNEI~\cite{devie2018intrinsic}, MATR~\cite{severson2019data, attia2020closed}, UL\_PUR~\cite{juarez2020degradation, juarez2021degradation}, SNL~\cite{preger2020degradation}, RWTH~\cite{li2021one}, and HUST~\cite{ma2022real}.  
These datasets span diverse electrode chemistries, cell formats, operating conditions, and degradation characteristics, capturing heterogeneity arising from both intrinsic cell properties and external cycling protocols. \cref{tab:dataset} summarizes their chemical compositions and cell-level statistics.
            
The MATR~\cite{severson2019data, attia2020closed} dataset comprises charge-discharge cycle data from 180 commercial LFP/graphite 18650 cylindrical cells (APR18650M1A, 1.1Ah, 3.3V nominal, A123 Systems). 
The dataset was collected for fast-charging optimization, and all experiments were conducted under a temperature-controlled environment of \SI{30}{\celsius}. All cells share identical discharge conditions, where discharge is performed using a 4C CC protocol down to 2.0V, while charging protocols are systematically varied. Charging consists of a fast-charging stage from 0-80\% SOC, followed by a uniform CC-CV stage from 80-100\% SOC.
We follow the original benchmark configuration, which partitions the dataset into MATR1, MATR2, and CLO.
In MATR1 and MATR2, multiple fast-charging policies are employed, categorized into one-step and two-step strategies for the 0-80\% SOC region. The CLO dataset, introduced by Attia \etal~\cite{attia2020closed}, further extends this design using a closed-loop optimization framework with multi-stage charging protocols.
MATR1 represents an in-distribution setting in which the training and test sets share identical charging policy distributions, with a stratified split to ensure balanced cycle-life coverage. In contrast, MATR2 evaluates generalization to previously unseen charging protocols, representing an out-of-distribution setting. CLO is constructed by randomly splitting the dataset without policy-based separation. Overall, the MATR dataset serves as a representative benchmark for analyzing the impact of charging-policy variations on early-cycle RUL prediction under controlled discharge conditions.

The HUST~\cite{ma2022real} dataset consists of 77 LFP/graphite cylindrical cells of the same type as the MATR dataset. 
In contrast to MATR, which emphasizes charging-policy diversity, HUST provides a complementary setting in which all cells follow a common fast-charging protocol but are subjected to diverse discharge strategies. 
All experiments were conducted in thermostatic chambers maintained at \SI{30}{\celsius}.
Charging follows a multi-stage CC-CV scheme, consisting of a high-rate CC stage at 5C from 0-80\% SOC, followed by a moderate CC stage at 1C until 3.6V, and a final CV stage at 3.6V with a current cutoff of C/20.
In contrast, discharging follows a four-stage constant-current scheme in which the current levels are systematically varied across the 100-60\%, 60-40\%, and 40-20\% SOC intervals, while the final stage from 20-0\% SOC remains fixed at 1C. Following the original experimental setup, we adopt the predefined split of 55 training and 22 test cells. This dataset emphasizes discharge-protocol diversity rather than charging-policy diversity.

The CALCE~\cite{xing2013ensemble, he2011prognostics} dataset includes full life cycle data from two groups of LCO/graphite prismatic cells: CS2 (1100mAh) and CX2 (1350mAh).
To isolate the effect of discharge conditions, all cells are charged under a standardized 0.5C CC-CV protocol up to 4.2V with a cutoff current of 0.05A, while the discharge conditions are intentionally varied.
The dataset includes diverse and non-ideal discharge behaviors such as variable-current profiles, random cut-off settings, partial cycling, pulsed discharge, and temperature-related variations. These variations reflect realistic and uncertain usage patterns, providing insight into model performance under diverse operating conditions.

The RWTH~\cite{li2021one} dataset comprises cycling data from 48 commercial NMC/graphite cylindrical cells (UR18650E, 1.1Ah, 3.7V nominal, Sanyo/Panasonic). All cells are aged under a common protocol at \SI{25}{\celsius}. The cycling procedure involves repeated CC-CV steps where charging is performed at 2C up to 3.9V and discharging proceeds down to 3.5V. Performance is monitored through initial characterization and periodic reference performance tests to track capacity fade and resistance evolution. This protocol provides a stable and controlled setting for analyzing degradation behavior under fixed operating conditions.
 
The SNL~\cite{preger2020degradation} dataset contains cycling data from 61 commercial 18650 cylindrical cells spanning three types of cathodes: 18 LFP (APR18650M1A, 1.1Ah, 3.3V nominal, A123 Systems), 22 NCA (NCR18650B, 3.2Ah, 3.6V nominal, Panasonic), and 21 NMC (18650HG2, 3.0Ah, 3.6V nominal, LG Chem), all paired with graphite anodes.
Unlike datasets with relatively fixed operating conditions, SNL systematically varies temperatures of \SI{15}{\celsius}, \SI{25}{\celsius}, and \SI{35}{\celsius}, along with discharge rates from 0.5C to 3C and SOC windows of 0-100\%, 20-80\%, and 40-60\%. While charging is consistently performed using a standardized CC-CV protocol at 0.5C up to the maximum voltage with a 0.05A cutoff, discharging is tailored to specific chemistry and SOC constraints. The dataset follows a structured, cycling process with periodic capacity checks, enabling the analysis of degradation behavior under various operating conditions.

The UL\_PUR~\cite{juarez2020degradation, juarez2021degradation} dataset comprises two experimental parts. In this study, we use Part 1, which contains commercial NCA/graphite cylindrical cells (NCR18650B, 3.4 Ah, 3.6 V nominal). All cells are cycled under a CC-CV protocol at near-ambient temperature of \SI{23}{\celsius}, and two voltage windows, 2.7-4.2V and 2.9-4.0V, are considered. Charging is performed at a constant current up to the upper cutoff voltage, followed by a constant-voltage phase until the current reaches a cutoff threshold, while discharging is conducted under constant-current operation down to the lower cutoff voltage. Unlike fixed-cycle protocols, aging is defined based on capacity-fade criteria rather than a predefined number of cycles. This setting provides a controlled framework for analyzing degradation under different operating limits.

\ExpDatasetFig

The HNEI~\cite{devie2018intrinsic} dataset consists of commercial 18650 cylindrical cells with blended LCO-NMC/graphite electrodes (ICR18650C2, 2.8Ah, 3.7V nominal, LG Chem). 
Following initial conditioning, a subset of cells is selected and aged under a fixed protocol at \SI{25}{\celsius}. The cycling protocol follows a standardized CC-CV scheme over a voltage range of 3.0-4.3V, involving charging at a C/2 constant current followed by a constant-voltage phase, while discharging at a constant current of 1.5C. The experiments consist of repeated 100-cycle blocks with periodic simplified reference performance tests, enabling analysis of intrinsic cell-to-cell variability under tightly controlled conditions.

Following the standardized preprocessing and deterministic train/test split protocols provided by the BatteryML framework~\cite{batteryml}, we construct eight benchmark datasets: MATR1, MATR2, HUST, SNL, CLO, CRUH, CRUSH, and MIX. Importantly, the BatteryML splitters operate at the cell-file level using predefined cell-ID lists rather than resampling train/test sets in each run, so that no cell appears in both the training and test sets, ensuring deterministic and reproducible evaluation.

CRUH aggregates multiple small-scale datasets, including CALCE, RWTH, UL\_PUR, and HNEI, each of which individually provides insufficient samples for stable model training. Integrating these datasets enhances diversity while retaining chemical and operational heterogeneity.

CRUSH extends CRUH by incorporating short-life cells from the SNL dataset, emphasizing early-cycle prediction scenarios with only a few initial cycles. This setting makes prediction more challenging by limiting observable degradation information and increasing variability across samples.

Finally, MIX represents the most comprehensive benchmark by integrating all datasets used in this study, thereby encompassing the highest level of data heterogeneity. As shown in \Cref{fig:dataset}, MIX exhibits significantly more diverse degradation trajectories and lifetime distributions compared to MATR1, reflecting the combined effects of variations in cell types, operating conditions, and degradation mechanisms. These characteristics make MIX the most challenging experimental setting for evaluating model robustness and generalization performance under diverse conditions.

\subsection{Evaluation metrics}
We evaluate the performance of the proposed battery RUL prediction method using three regression-based metrics commonly employed in RUL prediction tasks: root mean square error (RMSE), mean absolute error (MAE), and mean absolute percentage error (MAPE). These metrics measure the discrepancy between the predicted and actual EOL cycle, where EOL is defined as the cycle at which a battery capacity falls below a predefined SOH threshold. We adopt these three complementary metrics to provide a comprehensive evaluation of model performance, as battery RUL prediction requires not only minimizing overall error magnitude but also accounting for extreme prediction failures and relative error across different lifetime ranges.

RMSE computes the square root of the mean squared errors and emphasizes larger prediction errors due to the squared term:
\begin{equation}
RMSE = \sqrt{{{1}\over{n}}\sum^n_{i=1}(y_i-\hat{y_i})^2},
\label{eq:rmse}
\end{equation}
where $y_i$ and $\hat{y_i}$ denote the actual and predicted EOL cycles of the $i$-th battery cell, respectively, and $n$ is the total number of cells in the datasets. RMSE is a widely used standard metric for evaluating overall prediction accuracy, with lower values indicating better performance. In particular, it is sensitive to large errors, making it suitable for assessing robustness against extreme prediction failures.

MAE measures the average magnitude of prediction errors, regardless of whether the model overestimates or underestimates the actual value:
\begin{equation}
MAE = {{1}\over{n}} \sum^n_{i=1} \lvert y_i-\hat{y_i} \rvert,
\label{eq:mae}
\end{equation}
This metric is less sensitive to outliers compared to RMSE and provides a more interpretable measure of average error. As with RMSE, lower MAE values correspond to better prediction accuracy. Therefore, it reflects the overall stability and consistency of the model’s predictions across the entire lifetime range.

MAPE measures the average prediction error as a percentage of the actual value, providing a magnitude-agnostic evaluation of model performance:
\begin{equation}
MAPE = {{100\%}\over{n}} \sum^n_{i=1} \lvert {{y_i-\hat{y_i}}\over{y_i}} \rvert,
\label{eq:mape}
\end{equation}
This metric is particularly beneficial for comparing performance across datasets with varying lifetime scales. However, because MAPE is normalized by the true EOL, caution is required when interpreting MAPE alone, especially for samples with small true EOL values. Thus, RMSE, MAE, and MAPE are used as complementary metrics that capture absolute error magnitude, average cycle-level deviation, and relative error across lifetime scales, respectively.

\subsection{Implementation details}
We preprocess the data and train the model using the open-source BatteryML~\cite{batteryml} framework. BatteryML standardizes diverse public battery datasets into a unified format, provides benchmark-specific train/test cell splits, and defines a standardized labeling scheme for RUL prediction. 
In particular, the BatteryML splitters operate at the cell-file level using predefined cell-ID lists, ensuring deterministic and reproducible evaluation across different models.
The framework also adopts a modular design that separates preprocessing, feature extraction, and model training, enabling flexible modification and analysis at each stage. 

For generating time-frequency representations, we apply the CWT using the Morlet wavelet from the PyWavelets library~\cite{lee2019pywavelets}.
Following the scale-construction procedure described in \cref{subsec:wavelet_rul}, we use 32 scales in the 1-cycle setting and 256 scales in the 100-cycle setting. These original scalograms are then compressed to a common scale dimension of $r=16$ using group-wise averaging along the scale axis. The rationale for this compression level is analyzed quantitatively in \cref{subsec:redundancy}.
Model training is conducted in PyTorch~\cite{paszke2019pytorch} on a single NVIDIA RTX A5000 GPU, using ResNet18 as the CNN backbone with the Adam optimizer~\cite{kingma2014adam} and a learning rate of $10^{-3}$. 
To ensure reliable results, each experiment is conducted three times with different random seeds, and the average performance is reported. For a fair comparison, the baseline models keep their original hyperparameter settings whenever possible, with the number of input cycles adjusted to suit the specific experimental conditions.

\ExpRULLowCycleTab 

\subsection{Early-cycle RUL prediction}
\label{sec:early-cycle}
To evaluate the effectiveness of the proposed method, we conduct a comparative study using eight benchmark battery datasets against representative machine learning and deep learning approaches implemented based on the BatteryML~\cite{batteryml} framework. In particular, we focus on RUL prediction scenarios with severely limited early-cycle data, restricting the number of input cycles to a minimum. Furthermore, we analyze the robustness of prediction performance by gradually increasing the number of early cycles from 1 to 100, which is the standard setting commonly adopted in prior studies. 
In addition, we conduct further analyses to investigate the interpretability of the proposed method and its dependence on input signals. Specifically, we compare the performance across voltage, current, and capacity signals, and employ Grad-CAM-based visualization to identify the regions that the model attends to during prediction.

\cref{tab:rul_low_cycle} presents the RMSE results for RUL prediction when each method uses the minimum number of cycles it supports.
To evaluate performance under extremely limited data conditions, we compare the proposed method with a wide range of representative machine learning and deep learning models implemented based on the BatteryML~\cite{batteryml} framework, including Ridge regression, PCR, PLSR, Gaussian process, XGBoost, random forest, CNN, LSTM, MLP, and Transformer. In addition, recently proposed methods such as BatLiNet~\cite{zhang2025battery}, CPMLP, and CPTransformer~\cite{tan2025batterylife} are also included for comparison. All baseline models are implemented in their original configurations.
Most of these baseline methods rely on inter-cycle feature differences for prediction and therefore require at least two cycles of data.
Accordingly, the BatteryML-based machine learning and deep learning models, as well as BatLiNet, are evaluated using two cycles. In contrast, CPMLP, CPTransformer, and the proposed method can operate on a single-cycle input as they do not depend on inter-cycle variations.
Instead of relying on such inter-cycle variations, our method directly processes voltage, current, and capacity measurements from a single cycle to extract multi-resolution time-frequency features using the CWT. While recent models like CPMLP and CPTransformer also support single-cycle inputs as they do not necessitate features derived from cycle-to-cycle changes, our approach focuses on capturing intra-cycle dynamics through time-frequency analysis. 
This setting reflects the smallest feasible early-cycle input condition supported by each method while preserving the intended input structure of each baseline model.
To further enhance feature efficiency, we apply scale compression, which reduces redundancy in the high-dimensional representation while preserving features extracted from the time-frequency domain. As a result, the proposed method enables effective RUL prediction with only a single cycle of input data, without requiring a large number of cycles. 
Experimental results demonstrate that the proposed method, using only a single cycle, achieves the lowest RMSE on all eight benchmark datasets, with an average RMSE of 218.11. 

\ExpavgTab
\ExpHeatmapMetricsFig
The dataset-wise performance pattern is consistent with the characteristics of the benchmarks. On MATR1, where the training and test cells share matched charging-policy distributions under a common discharge protocol, the proposed method reduces the RMSE from 139.52 to 101.69, indicating that the time-frequency representation effectively captures subtle early-cycle degradation patterns under controlled conditions. The improvement remains substantial on MATR2, where the RMSE decreases from 352.60 to 290.58, and on CLO, where the RMSE decreases from 123.04 to 102.63 despite the more complex temporal dynamics introduced by multi-stage closed-loop fast charging. In contrast, the performance margins are smaller on HUST, SNL, CRUH, CRUSH, and MIX, where prediction is made more difficult by discharge-protocol diversity, mixed chemistries, short-life cells, or cross-dataset heterogeneity. Nevertheless, the proposed method still achieves the best RMSE on these benchmarks, including MIX, the most heterogeneous setting in this study, with an RMSE of 179.68. 
These results suggest that the proposed approach is effective for extracting lifetime-relevant signal patterns while maintaining robust performance across chemically and operationally diverse public benchmark datasets.

This dataset-wise tendency is also reflected in the aggregate results summarized in \cref{tab:avg_eval_table}. Despite using only the first cycle as input, the proposed method achieves the lowest average RMSE, MAE, and MAPE across all eight benchmark datasets, with values of 218.11, 144.02, and 0.20, respectively. Compared with BatLiNet, which requires two cycles, the proposed method reduces the average RMSE from 281.79 to 218.11, corresponding to a 22.6\% improvement. Relative to the recent single-cycle baselines CPMLP and CPTransformer, the RMSE reductions are 21.5\% and 18.7\%, respectively. The concurrent improvements in MAE and MAPE indicate that the proposed representation not only suppresses large prediction failures but also reduces the average absolute and relative prediction errors across datasets with different lifetime ranges. Overall, these results confirm that the proposed method provides balanced gains across the three complementary evaluation metrics under the most data-constrained setting.

\Exppredictplots
We additionally evaluate the robustness of our method to variations in the number of early input cycles. 
\Cref{fig:heatmap_metrics} presents a heatmap of the average RMSE (\Cref{fig:hm_rmse}), MAE (\Cref{fig:hm_mae}), and MAPE (\Cref{fig:hm_mape}) across eight benchmark datasets, where the number of input cycles is gradually increased from 1 to 100. Since the Full, Discharge, and Variance models~\cite{severson2019data} require at least 5 cycles, whereas the most conventional machine learning and deep learning models~\cite{batteryml} and BatLiNet~\cite{zhang2025battery} require at least 2 cycles, the experiments are conducted accordingly. Regions with fewer cycles than required are marked in white to indicate missing values. In contrast, because our method and recent architectures such as CPMLP and CPTransformer~\cite{tan2025batterylife} can operate on a single cycle, their performance can be evaluated even under severely data-limited conditions. Existing approaches often suffer substantial performance degradation when fewer cycles are available and typically require a sufficient number of cycles to maintain low prediction errors. By contrast, our method consistently maintains low prediction errors across all input-cycle conditions, demonstrating strong robustness even in severely data-limited scenarios.

In addition, \Cref{fig:mix_prediction_all} presents a visual comparison between our method and BatLiNet~\cite{zhang2025battery} on the MIX dataset under different input cycle conditions. When BatLiNet uses 2 cycles (\Cref{subfig:batlinet_2}) and our method uses 1 cycle (\Cref{subfig:ours_1}), our predictions show a noticeably tighter alignment with the ground-truth diagonal. Likewise, under the 100-cycle setting, our method (\Cref{subfig:ours_100}) again outperforms BatLiNet (\Cref{subfig:batlinet_100}), producing less scatter and more accurate predictions across the full range of ground-truth values. 
These findings suggest that the proposed wavelet-based feature extraction approach is robust to variations in input cycle count and remains effective in data-limited scenarios, while providing informative feature representations for deep learning models without additional architectural modifications.
\Expgradcam
\Expviqdepend

We further investigate whether the proposed model overly relies on a specific input channel by comparing RUL prediction performance under different input signal configurations, using voltage, current, and capacity either individually or in combination. The experiment is conducted on the MIX dataset, and the results are summarized in \cref{tab:viq_dependency}. All configurations follow the same feature extraction pipeline: each signal is independently transformed into a time-frequency representation using the CWT and then compressed along the scale axis to form the model input.
As shown in \cref{tab:viq_dependency}, the full three-channel configuration (VIC) achieves the best performance across all metrics, indicating that voltage, current, and capacity provide complementary degradation information. Among the single-channel settings, current and voltage achieve relatively stable performance, whereas the capacity-only setting results in the largest error. Notably, pairwise combinations do not consistently outperform the best single-channel setting, suggesting that simply adding one additional signal does not necessarily improve prediction accuracy.
The competitive performance of the voltage-only and current-only settings indicates that the proposed representation does not overly depend on a specific input channel. In contrast, the relatively poorer performance of the capacity-only setting suggests that capacity changes during early cycles are too limited to provide sufficiently informative degradation cues.
Overall, these findings support the effectiveness of multi-signal fusion and suggest that capacity provides complementary information when used together with voltage and current, while voltage and current signals also contain useful predictive cues under controlled benchmark conditions.

To gain insight into the time-frequency patterns utilized by the model, we conduct a Grad-CAM-based explainability analysis. \Cref{fig:gradcam_three_channels} presents the visualization results for a cell (MATR\_b1c1) included in the MIX dataset, using voltage, current, and capacity as input channels.
Each column corresponds to a different input channel; the first row shows the original signal, the second row shows the corresponding CWT scalogram, and the third row shows the Grad-CAM activation map associated with that scalogram. The activations are not uniformly distributed over the entire scalogram, but are instead concentrated in localized temporal regions, mainly at lower scales. Because lower scales in the CWT correspond to higher-frequency components, this pattern suggests that the model primarily relies on localized morphological transitions and short-duration structural changes rather than on globally averaged signal magnitude alone. More specifically, in the voltage channel, relatively strong activation appears around regions where the slope of the charge profile changes, such as plateau-transition-like segments. In the current channel, attention is focused near switching regions between the charge, rest, and discharge regimes. In the capacity channel, activation appears around regions where the local rate of capacity accumulation changes. 
These channel-dependent activation patterns indicate that different input signals emphasize distinct time-scale structures associated with the charge-discharge operation. The Grad-CAM shows that the trained model attends to localized structural variations in the CWT scalograms during prediction, suggesting that these regions contribute useful signal-level cues for RUL prediction. These qualitative findings complement the input-signal ablation results by showing that the model can learn channel-specific informative patterns directly from the data.

\subsection{Effect of time-frequency feature extraction and scale compression}
\label{subsec:cwt}
\Exptimefrequency
In this section, we examine whether time-frequency-based feature extraction provides advantages over direct time-domain modeling for battery degradation signals, and whether scale compression further improves both prediction accuracy and computational efficiency. We also include an auxiliary knee-point prediction analysis to examine whether the proposed representation captures degradation-related structure beyond EOL prediction.

To validate the choice of CWT, we compare it with two representative alternatives, STFT and DWT. 
For STFT, we use a Hann-window-based spectrogram, whereas for DWT, we use the db4 mother wavelet with decomposition level 4 and construct features from the approximation and detail coefficients. \cref{tab:timefrequency} reports the average RUL prediction performance of these methods across all benchmark datasets under the 1-cycle setting. The results show that DWT and CWT outperform the raw time-domain signal across all three metrics, while STFT improves RMSE and MAE but yields a slightly higher MAPE. Among the compared methods, CWT achieves the best overall performance, with an RMSE of 227.7, MAE of 152.0, and MAPE of 22.8\%. 
STFT relies on a fixed window, whereas DWT is based on a discrete hierarchical scale structure. These characteristics may limit how flexibly they represent local temporal patterns across scales in the single-cycle battery signals considered here. By contrast, CWT provides a continuous scale representation, which better matches the weak and non-stationary signal-level variations available from a single cycle in our experiments. Based on this task-specific comparison, we adopt the CWT-based time-frequency representation as the primary feature extraction method in the proposed framework.
This observed advantage is interpreted within the single-cycle battery prediction setting considered here, rather than as a general claim of superiority.

\ExpwwoCWTFig
We conduct an ablation study to evaluate the effectiveness of CWT-based feature extraction and the incremental effect of scale compression. Specifically, we compare three input configurations: (1) a time-domain representation obtained by linearly interpolating the voltage, current, and capacity signals as defined in \cref{eq:raw_extract}; (2) a CWT-based time-frequency representation extracted from the same interpolated inputs; and (3) a compressed CWT representation obtained by compressing the CWT output along the scale axis. Each configuration is evaluated across all benchmark datasets under two conditions: 1 cycle (minimal input) and 100 cycles (substantially richer input). All experiments utilize the same CNN architecture, with only the input channels modified according to each configuration. \Cref{fig:wwoCWT} presents the average RMSE, MAE, and MAPE across all benchmark datasets. Applying the CWT to time-domain inputs improves RMSE and MAE in both input settings.
Specifically, in the 1-cycle setting, RMSE decreases from 274.3 to 227.7, and MAE from 192.5 to 152.0, and MAPE from 26.9\% to 22.8\%. In the 100-cycle setting, RMSE decreases from 255.7 to 219.1, and MAE from 166.5 to 151.4, whereas MAPE increases from 23.8\% to 29.2\%. When scale compression is additionally applied, the compressed CWT representation achieves the lowest prediction errors across all three metrics in both input settings.
In the 1-cycle setting, RMSE decreases further to 218.1, MAE to 144.0, MAPE to 20.1\%. For 100 cycles, the compressed representation yields an RMSE of 210.7, an MAE of 139.1, and a MAPE of 21.4\%.

\ExpTrainingTime
To analyze the effect of scale compression on training efficiency, we compare model complexity and training cost with and without compression. \cref{tab:training_time2} reports the number of parameters, the per-sample GFLOPs, and the total training time under the 1-cycle and 100-cycle settings. Because the original CWT-based representation contains a dense multi-scale dimension, using it without compression increases the effective input dimensionality and the computational burden of the downstream network. In the 1-cycle setting, the number of scales is reduced from 32 to 16, whereas in the 100-cycle setting, it is reduced from 256 to 16. Accordingly, scale compression reduces the scale dimensionality, leading to fewer parameters, lower computational cost, and shorter training time. Under the 1-cycle setting, the parameter count decreases from 11.47M to 11.32M, the GFLOPs from 0.40 to 0.36, and the total training time from 313.48 s to 225.84 s. Similarly, in the 100-cycle setting, the parameter count decreases from 13.73M to 11.32M, the GFLOPs from 38.43 to 4.56, and the training time from 1522.05 s to 240.06 s. 
These results show that compressing the scale dimension reduces the computational burden associated with CWT-based representations during model training.
\Expkneepoint

We also include an auxiliary knee-point prediction analysis as a data-driven check of whether the proposed single-cycle representation contains predictive cues related to the knee point, another degradation-related trajectory target beyond EOL prediction.
The knee point denotes the point on the capacity degradation curve at which the rate of capacity fade changes abruptly, and it is widely used as an informative descriptor of battery aging behavior. Such knee-point analysis has been widely studied in the context of battery degradation characterization and prediction~\cite{zhang2019accelerated, fermin2020identification, wang2024capacity}. Following Ferm\'{\i}n-Cueto \etal~\cite{fermin2020identification}, we fit each capacity degradation curve using the Bacon-Watts (BW)~\cite{bacon1971estimating} model and derive the knee-point label from the fitted trajectory.

Unlike RUL labels, which are typically defined using a fixed EOL threshold, knee-point labels depend on the shape of the degradation curve and are therefore more sensitive to dataset-specific noise. As shown in \Cref{fig:knee_a}, the predicted knee is closely aligned with the fitted label for the smoother MATR example. By contrast, in the SNL example shown in \Cref{fig:knee_b}, repeated spikes make the BW-based label more ambiguous and the prediction substantially more difficult. This highlights an inherent limitation of knee-point definition on noisy real-world datasets.
Accordingly, knee-point prediction performance varies substantially across benchmarks. Under the same 1-cycle setting used for RUL prediction, the proposed method yields lower RMSE values on smoother benchmarks such as MATR1 and CLO, whereas noisier and more heterogeneous datasets such as SNL and CRUSH exhibit substantially larger errors, as shown in \Cref{fig:knee_c}.
These results provide preliminary evidence that the proposed single-cycle representation may contain predictive cues related to degradation-trajectory targets beyond EOL, although the strong dataset dependence of knee-point prediction suggests that this auxiliary task remains more challenging than EOL prediction.

\subsection{Scale compression via redundancy-fidelity analysis}
\label{subsec:redundancy}
To characterize the redundancy-fidelity trade-off introduced by scale compression and to select a practical operating compression level, we analyze the compressed scalogram representation across different compression levels.
Here, the original scalogram ${S}_x$ and the compressed scalogram $\tilde{S}_x$ denote the CWT-based time-frequency representation defined above and its compressed version obtained through scale-axis compression, respectively. These two criteria are used as complementary measures to assess how much redundant information remains within the compressed representation and how well the original time-frequency structure is preserved during compression.

We first define internal redundancy as the degree to which different scale vectors in the compressed scalogram contain similar information. Since the CWT is inherently overcomplete, strong correlations may arise between adjacent scales. Indeed, when we analyze the inter-scale correlation structure of the original scalogram, the scale-scale correlation matrix exhibits a pronounced diagonal band pattern, indicating that neighboring scales tend to share similar frequency bands and therefore contain highly similar information. To quantify this redundancy in the compressed representation, we extract vectors along the scale axis from each channel-specific compressed scalogram $\tilde{S}_i^{(h)}$ and construct a scale-vector matrix $Z \in \mathbb{R}^{r \times T}$, where $r$ denotes the number of compressed scales and $T$ is the temporal length. Each row vector represents the time-frequency response at one compressed scale. We then analyze the covariance structure of $Z$ to quantify redundancy across scales. Let $\lambda_1, \ldots, \lambda_r$ denote the eigenvalues of the covariance matrix $\mathrm{cov}(Z)$. The normalized eigenvalue distribution is defined as $p_i = \lambda_i / \sum_{j=1}^{r} \lambda_j$. Based on this distribution, the effective rank~\cite{roy2007effective}, which represents the effective dimensionality of the scale representation, is computed as follows:
\begin{equation}
r_{\mathrm{eff}}(Z)=\exp
\left(
-\sum_{i=1}^{r} p_i \log p_i
\right).
\label{eq:effectiverank}
\end{equation}

The effective rank indicates the intrinsic dimensionality occupied by the compressed scale representation; it becomes smaller when information across scales is concentrated in a low-dimensional subspace. By normalizing this quantity with respect to the total number of scales, we define the redundancy ratio as:
\begin{equation}
RR(Z) = 1- \frac{r_{\mathrm{eff}}(Z)}{r},
\label{eq:redundancy_ratio}
\end{equation}
where a larger value indicates greater inter-scale redundancy. In this work, we use the effective-rank-based redundancy ratio as the primary redundancy measure for evaluating different levels of scale compression.

To assess the information loss introduced by scale compression, we further define a fidelity metric that measures how well the compressed representation preserves the original time-frequency structure. For this purpose, we perform a block-wise correlation analysis that evaluates how well each compressed scale represents its corresponding block of original scales. Specifically, we partition the original scale set into adjacent scale blocks $G_g$. Let $\tilde{s}_g$ denote the compressed scale vector extracted from the compressed scalogram, and let $s_a$ denote an original scale vector belonging to the corresponding block. The block-wise fidelity is then defined as:
\begin{equation}
f_g = \frac{1}{|G_g|}\sum_{a \in G_g}
\left|
\mathrm{corr}(\tilde{s}_g, s_a)
\right|,
\label{eq:blockwise_corrleation}
\end{equation}
which reflects how well the compressed scale represents the time-frequency structure of the corresponding original scale block. The overall fidelity is computed by averaging over all scale blocks:
\begin{equation}
Fid=\frac{1}{r}\sum_{g=1}^{r}f_g.
\label{eq:fidelity}
\end{equation}
In the experiments, the same computation is performed for the voltage, current, and capacity channels, and the final fidelity score is obtained by averaging across channels. In addition, to complement the correlation-based fidelity analysis, we also performed auxiliary evaluations based on cosine similarity, mutual information, and block-wise KL divergence, all of which exhibited trends consistent with the correlation-based results.

As the degree of scale compression becomes stronger, internal redundancy decreases, but fidelity to the original scalogram also tends to decrease. 
Therefore, a practical compression level can be determined by balancing redundancy reduction against structural preservation. We use the following operating criterion:
\begin{equation}
r_{\mathrm{op}}=\arg\min_r RR(Z_r)\quad\text{s.t.}\quad Fid(Z_r) \ge \tau.
\label{eq:optimalselection}
\end{equation}
Here, $r_{\mathrm{op}}$ denotes the practical operating compression level determined among the evaluated candidate compression levels in each setting, while $RR(Z_r)$ and $Fid(Z_r)$ denote the redundancy ratio and the block-wise correlation-based fidelity at compression level $r$, respectively. 
The threshold $\tau$ restricts consideration to candidates that retain a correlation-based fidelity of at least $\tau$ to the original scalogram.
\Expredundancy
\Cref{fig:redundancy_fidelity} illustrates the trade-off between redundancy and fidelity under different compression levels. To examine how scale compression behaves under different time-frequency resolutions, we consider two CWT settings. \Cref{fig:compression_small} shows the results for a lower-resolution scalogram with 32 scales compressed under the candidate settings $r = 2, 4, 8, 16$, whereas \Cref{fig:compression_large} presents the results for a higher-resolution scalogram with 256 scales under $r = 8, 16, 32, 64$. As shown in the figure, stronger compression is effective in reducing inter-scale redundancy, but it also lowers fidelity to the original scalogram. This trade-off indicates that overly aggressive scale compression can remove redundant information efficiently, while simultaneously discarding part of the original time-frequency structure. Therefore, it is important to choose an appropriate compression level by balancing redundancy reduction against fidelity preservation. 
The fidelity threshold defines the admissible candidate range, so changing $\tau$ (e.g., to $0.90$ or $0.99$) can alter the resulting operating level. We use $\tau = 0.95$ as a practical high-fidelity compromise between redundancy reduction and fidelity preservation; under this threshold, \cref{eq:optimalselection} yields $r_{\mathrm{op}} = 16$ as the common operating level for both scale settings. This operating level also coincides with the compression level yielding the lowest RMSE in both \Cref{fig:compression_small} and \Cref{fig:compression_large}.

\ExpCompressMethodTab

\subsection{Comparison of the CWT scale compression methods}
\label{subsec:ablation}
Time-frequency representations based on the CWT can capture diverse signal characteristics across multiple scales. However, this multi-resolution property also increases the input dimensionality, resulting in higher computational cost and potential redundancy in the extracted features. To address these issues, we compare several scale compression methods designed to reduce inter-scale redundancy while preserving informative time-frequency components.

\cref{tab:compression_method} presents a comparative evaluation of five scale compression methods under both the 1-cycle and 100-cycle settings. Principal component analysis (PCA), a classical statistical approach, computes principal components across all scales and projects the original high-dimensional coefficients onto a lower-dimensional subspace of $r$ components along directions of maximal variance. The top-$k$ energy method selects the $k$ scales with the highest energy, where $k$ is set to $r$, based on the assumption that higher-energy scales contain more relevant information. However, such global reduction strategies may discard important localized time-frequency patterns that are useful for battery degradation modeling. To better reflect the local structure of the CWT representation, we further consider group-based dimensionality reduction methods motivated by the observation that adjacent scales often exhibit strong redundancy, as described in \cref{eq:scalecompression}.
The energy-weighted method computes the energy of all scales, normalizes the energy values, and then groups adjacent scales. Within each group, a weighted average is calculated using the normalized energy as weights. This approach aims to emphasize contributions from high-energy scales while suppressing those from lower-energy scales. The group-max and group-mean methods use the same grouping strategy, but reduce dimensionality by taking either the maximum or the mean value within each group. These methods produce compact representations while preserving local time-frequency characteristics and mitigating redundancy in battery degradation signals.

As shown in \cref{tab:compression_method}, the group-mean method consistently achieves the best prediction performance across all evaluation metrics in both cycle settings. In the 1-cycle setting, it attains the lowest RMSE, MAE, and MAPE of 218.11, 144.02, and 0.20, respectively, and it remains the best-performing method in the 100-cycle setting with values of 210.72, 139.07, and 0.21. 
By contrast, methods that emphasize only a subset of dominant scales, such as top-$k$ energy and group-max, yield inferior results, suggesting that selectively retaining only the strongest responses is insufficient for representing the full scale distribution relevant to degradation dynamics. Although the energy-weighted method performs better than the global baselines, it still underperforms the group-mean approach, indicating that overemphasizing high-energy scales may bias the representation toward limited components of the signal. Overall, the group-mean operation preserves a more balanced summary of the scale distribution and leads to more stable and accurate RUL prediction.
\Expablation

To further characterize the scale compression strategies, we additionally compare these strategies using the redundancy and fidelity measures defined in \cref{subsec:redundancy}. 

\Cref{fig:ablation_fid} shows the relationship between internal redundancy and fidelity to the original scalogram for the different compression strategies. Among the globally applied compression methods, PCA reduces internal redundancy the most, but it exhibits the lowest fidelity, indicating that it fails to preserve the original time-frequency structure sufficiently during compression. The top-$k$ energy method shows higher fidelity than PCA, but it also yields the highest redundancy, implying that redundant inter-scale information is not effectively removed. These results suggest that selecting or compressing scales solely according to global variance or energy criteria may not adequately reflect the structural characteristics of the time-frequency representation.
In contrast, the group-based methods generally preserve higher fidelity to the original representation. The energy-weighted method maintains high fidelity, but still exhibits relatively large redundancy, which may arise because weighting high-energy components within each scale group repeatedly emphasizes similar scale information. The group-max method reduces redundancy to a level comparable to that of the group-mean method, but its fidelity is lower, likely because taking only the maximum response within each group selectively amplifies limited time-frequency components while suppressing the broader structural distribution. By comparison, the group-mean method achieves the most balanced behavior, maintaining high fidelity while effectively reducing internal redundancy. This suggests that averaging within each scale group alleviates redundant information from neighboring scales while preserving the overall distribution of the original time-frequency structure in a stable manner.

Notably, this redundancy-fidelity analysis is consistent with the RUL prediction results in \cref{tab:compression_method}: the group-mean method yields the best RMSE, MAE, and MAPE, while also providing the most favorable balance between information preservation and redundancy reduction. Therefore, considering not only prediction accuracy but also the structural properties of the compressed scale representation, we adopt the group-mean method as the final scale compression strategy in the proposed framework.

\section{Discussion}
This study shows that the proposed CWT-based representation with scale compression is effective for early-cycle battery RUL prediction, particularly in the single-cycle setting where only a highly limited early-life observation is available. Despite these advantages, several limitations remain.
First, although scale compression reduces the dimensionality of the CWT scalogram and improves the efficiency of the downstream prediction model, it does not reduce the computational cost of the CWT itself. Because the proposed framework first generates the original CWT representation and then applies compression, the computational burden of the wavelet transform remains unchanged. This limitation is less critical in the 1-cycle setting, where the input length is short, but becomes more evident as the number of input cycles increases.

Second, although the proposed method was evaluated on multiple public datasets with heterogeneous chemistries, form factors, and cycling protocols, the experiments were still conducted mainly under controlled laboratory cycling conditions. These datasets provide a standardized basis for evaluating early-cycle prediction performance across diverse battery domains, but they do not fully capture practical EV or energy-storage operating conditions, such as variable temperatures, partial cycling, irregular load profiles, changing C-rates, and diverse usage histories. Therefore, the current results should be interpreted as demonstrating robustness across controlled public benchmark settings, while further validation using real operating data is required to assess applicability under practical deployment conditions.

These limitations also suggest several directions for future research. Improving the efficiency of the time-frequency transformation itself remains an important priority. 
In particular, it would be meaningful to investigate fast CWT variants such as fCWT~\cite{arts2022fast} or to integrate learnable wavelet-based modules into the prediction network, thereby retaining the benefits of multi-resolution analysis with lower computational overhead. In addition, because prediction performance is influenced by dataset- and cycle-dependent scale configurations, the development of adaptive scale-selection or compression strategies may further enhance the framework's flexibility. 
Finally, future work may benefit from architectures that explicitly incorporate auxiliary condition data, such as temperature, charging conditions, operating profiles, C-rate variations, and partial-cycling information. Evaluating such models using real operating logs, together with uncertainty-aware evaluation and representative failure-mode analysis, would also be useful for assessing their robustness under practical EV and energy-storage operating conditions.

\section{Conclusion}
In this study, we proposed a framework for battery remaining useful life prediction designed for the single-cycle setting.
By transforming early-cycle battery signals into time-frequency representations using the continuous wavelet transform with scale compression, the proposed framework provides an integrated representation pipeline for single-cycle RUL prediction while reducing redundancy in dense scalograms.
Extensive experiments on eight benchmark datasets derived from seven public battery datasets showed that the proposed framework achieves strong prediction performance under the 1-cycle setting and remains robust as the number of input cycles increases from 1 to 100. Additional analyses further supported the effectiveness of time-frequency representations over direct time-domain input and showed that scale compression provides a favorable balance between redundancy reduction and fidelity preservation.

These results demonstrate that time-frequency-based multi-resolution analysis can serve as an effective and interpretable feature-extraction strategy for early-cycle battery remaining useful life prediction. At the same time, this study has several limitations. Although scale compression improves the efficiency of the downstream prediction model, it does not reduce the computational cost of the continuous wavelet transform itself, and the experiments were conducted mainly under controlled laboratory conditions. Future work will therefore investigate more efficient time-frequency analysis schemes, adaptive scale-selection or compression strategies, and architectures that incorporate auxiliary metadata such as temperature and operating conditions. Overall, the proposed method could serve as a promising approach for early battery lifetime evaluation, helping reduce the time and resources required for battery development and validation, while also providing a useful basis for incorporating time-frequency representations into advanced models for battery health prediction.

\appendix

\section{Data availability}
All datasets used in this study are publicly available through MATR~\cite{matr_dataset}, HUST~\cite{hust_dataset}, CALCE~\cite{calce_dataset}, and RWTH~\cite{rwth_dataset}. The SNL, UL\_PUR, and HNEI datasets are accessible through the BatteryArchive~\cite{battery_archive}.

\bibliographystyle{elsarticle-num}

\bibliography{cas-refs}

\end{document}